\documentclass[sigconf,nonacm]{aamas} 

\usepackage{balance} 

\usepackage{multirow}%
\usepackage{xcolor}%
\usepackage{tabularx}
\usepackage{makecell}
\usepackage{ragged2e}
\usepackage{enumitem}

\usepackage{color}
\usepackage{dsfont}
\usepackage{algorithm}
\usepackage{orcidlink}
\usepackage{algpseudocode}
\usepackage{utfsym}
\usepackage{makecell}
\usepackage{bbm}

\title{Learning the Value Systems of Societies with Preference-based Multi-objective Reinforcement Learning}

\author{Andrés Holgado-Sánchez\orcidlink{0000-0001-8853-1022}}
\affiliation{
  \institution{CETINIA, University Rey Juan Carlos}
  \city{Madrid}
  \country{Spain}}
\email{andres.holgado@urjc.es}

\author{Peter Vamplew\orcidlink{0000-0002-8687-4424}}
\affiliation{
  \institution{Federation University}
  \city{Ballarat}
  \country{Australia}}
\email{p.vamplew@federation.edu.au}

\author{Richard Dazeley\orcidlink{0000-0002-6199-9685}}
\affiliation{
  \institution{Deakin University}
  \city{Geelong}
  \country{Australia}}
\email{richard.dazeley@deakin.edu.au}

\author{Sascha Ossowski\orcidlink{0000-0003-2483-9508}}
\affiliation{
  \institution{CETINIA, University Rey Juan Carlos}
  \city{Madrid}
  \country{Spain}}
\email{sascha.ossowski@urjc.es}

\author{Holger Billhardt\orcidlink{0000-0001-8298-4178}}
\affiliation{
  \institution{CETINIA, University Rey Juan Carlos}
  \city{Madrid}
  \country{Spain}}
\email{holger.billhardt@urjc.es}

\begin{abstract}

Value-aware AI should recognise human values and adapt to the value systems (value-based preferences) of different users. This requires operationalization of values, which can be prone to misspecification. The social nature of values demands their representation to adhere to multiple users while value systems are diverse, yet exhibit patterns among groups. In sequential decision making, efforts have been made towards personalization for different goals or values from demonstrations of diverse agents. However, these approaches demand manually designed features or lack value-based interpretability and/or adaptability to diverse user preferences.

We propose algorithms for learning models of value alignment and value systems for a society of agents in Markov Decision Processes (MDPs), based on clustering and preference-based multi-objective reinforcement learning (PbMORL). We jointly learn socially-derived value alignment models (groundings) and a set of value systems that concisely represent different groups of users (clusters) in a society. Each cluster consists of a value system representing the value-based preferences of its members \highlight{and an approximately Pareto-optimal policy that reflects behaviours aligned with this value system. We evaluate our method against a state-of-the-art PbMORL algorithm and baselines on two MDPs with human values.}
\end{abstract}


\newcommand{\BibTeX}{\rm B\kern-.05em{\sc i\kern-.025em b}\kern-.08em\TeX}

\newcommand{\R}{\mathbbm{R}}

\newcommand{\Rv}{{\pmb{R}}}
\newcommand{\A}{\mathcal{A}}
\newcommand{\T}{\mathcal{T}}

\newcommand{\abs}[1]{{\left|#1\right|}}
\DeclareMathOperator*{\argmax}{arg\,max}
\DeclareMathOperator*{\argmin}{arg\,min}

\newcolumntype{Y}{>{\Centering\arraybackslash}X}

\newcommand{\highlight}[1]{{{#1}}}
\newcommand{\conc}{{\textproc{conc}}}
\newcommand{\chr}{\textproc{chr}}
\newcommand{\repr}{\textproc{repr}}
\newcommand{\VS}{\text{VS}}

\begin{document}


\pagestyle{fancy}
\fancyhead{}


\maketitle 

\sloppy
\section{Introduction}\label{sec:introduction}

To remain human-aligned, AI agents require awareness of human values such as \textit{benevolence} or \textit{achievement}~\cite{valueengineeringAutonomous2023}, i.e. the ability to reason with and about the value-alignment of different actions. This allows making explicit the alignment with the \textit{pluralistic} \textit{value systems}~\cite{leraleri2024aggregation} of human agents, manifested as value-based preferences\footnote{Here, ``value-based preferences'', refer to preferences between alternatives based on their alignment~\cite{Russell2022alignmentDefinition} with multiple values.}.

Value-awareness requires computational representations of values: existing works model human values with, e.g. utility functions~\cite{montes2022synthesis,Serramia2018}, reward functions~\cite{rodriguez2026reinforcementEthicalEmbedingWeightsRewardRL,wang-etal-2024-interpretableRewardModelingLLM}, constraints~\cite{zhong2024provablemultipartyreinforcementlearning,Balakrishnan2019behaviouralalignment} or logic programming~\cite{Anderson2018}. Value systems are typically represented by value orderings~\cite{serramia2023EncodingValueAlignedNorms,rodriguez2026reinforcementEthicalEmbedingWeightsRewardRL}, weights over value alignment utilities/rewards~\cite{AaronAgreementTOPSIS,leraleri2024aggregation},  preferences over alternatives~\cite{Chaput2023CONTEXTUALpreferenceToSettleDilemmas, andresEcai2025} or other methods such as taxonomies~\cite{Osman2024}. Such models can be applied to choosing value-aligned government policies~\cite{AaronAgreementTOPSIS}, obtaining aligned behaviour~\cite{Peschl2022LearnPreferencesFromExperts,rodriguez2026reinforcementEthicalEmbedingWeightsRewardRL} or norm selection in multi-agent systems~\cite{aguilera2024poverty,montes2022synthesis}.

However, representing values is challenging, due to their context-sensitive~\cite{Liscio2022Axies2}, and socially emergent and evolving nature~\cite{schwartz1992universals,Osman2024}. This often results in misspecification~\cite{Sumers2022InstructionsAndDescriptions}. Also, most of the surveyed works that analyse value-based preference diversity (e.g.~\cite{AaronAgreementTOPSIS,leraleri2024aggregation}) cannot simultaneously represent the various levels of alignment with values demanded by different user cohorts~\cite{Wang2024MAP:Palette}. 

For Markov Decision Processes (MDPs), reward learning is a promising way to achieve value-alignment~\cite{virginiaDignumResponsibleAutonomy2017,Leike2018ScalableAA}. Many works, though, focus on human alignment for the whole society~\cite{dpoLLM2023}, \highlight{but without awareness of the multiple value-based preferences involved. As such, we argue that value alignment should be treated as a multi-objective problem instead~\cite{Vamplew2018Human-alignedProblem}; in particular, encoding} separate values as different components of a reward vector allows an AI system to recognise these values and abide with them under varying conditions. Modelling values and value-based preferences of specific users has been approached via multi-objective rewards~\cite{rodriguez2026reinforcementEthicalEmbedingWeightsRewardRL,yangRewardsInContextMOAlignment,wang-etal-2024-interpretableRewardModelingLLM}, but these works struggle to instantiate models that concisely represent the diversity of value preferences within human groups.

\highlight{In this paper, we propose an online preference-based multi-objective reinforcement learning~\cite{Mu2024Preference-basedModeling} algorithm with clustering that learns the value system of a society in a MDP. This entails learning: a) a social \textit{grounding} (reward vector) that represents the alignment of different behaviours with a given set of values; b) a set of values systems (linear scalarization functions) that represent different clusters of agents with similar value-based preferences; and c) a set of MDP policies, each aligned with one of those value systems. We set out from a \textit{social value system learning} algorithm ~\cite{andresEcai2025} that learns a) and b) in non-sequential decision-making. Our algorithm asks for pairwise preferences between trajectories \highlight{(i.e. sequences of observations and actions)} in terms of the value systems of each agent (value-based preferences) and in terms of their alignment with the given set of values (value alignment preferences). We evaluate the capabilities of the method to approximate the value systems and behaviours of societies in synthetic MDPs.}



The paper is structured as follows. In Section~\ref{sec:background-valuealignment} we propose a representation of the value system of a society. In Section~\ref{sec:proposal} we describe the social value system learning problem and propose an algorithm to solve it. In Section~\ref{sec:evaluation} we evaluate our methods against baselines and an existing PbMORL algorithm~\cite{Mu2024Preference-basedModeling}. In Section~\ref{sec:conclusions}, we provide conclusions, limitations, and suggestions for future work.


{\section{Related works}\label{sec:state-of-art}

\subsection{Value Awareness, Inference and Learning}

Artificial Moral Agents (AMAs)~\cite{wallach2008moral} can be classified as explicit or implicit depending on whether their behaviour can be expressed in ethical terms~\cite{Cervantes2020}. Implicit systems acquire \textit{(value-)aligned} behaviour through imitation~\cite{dpoLLM2023,Pommeranz2012} or teaching~\cite{hadfieldCIRL,Malik2018CIRLimproved}; explicit ones operationalize values or principles to permit the ethical evaluation of actions~\cite{Anderson2018} or norms~\cite{Neufeld202425,Neufeld2022444}.
Value-aware systems~\cite{valueengineeringAutonomous2023} extend AMAs by explicitly grounding values as desirable goals~\cite{schwartz2012overview} and respecting agents’ \textit{value systems} (preferences over values~\cite{Rohan2000}).


Given the difficulty of eliciting values and value systems, Liscio et al.~\cite{Liscio2023Sociotechnical} propose \emph{value inference} as a three-step process, namely: i) \emph{value identification}~\cite{Pommeranz2012,Liscio2022Axies2}, which finds the values relevant to a society; ii) \emph{value system estimation}~\cite{valuesystemestimationAraque2020,Siebert2022liscio}, which learns the value systems of its members; and iii) \emph{value system aggregation}~\cite{AaronAgreementTOPSIS,leraleri2024aggregation}, which derives a value system to represent the whole society. 



Value inference, however, does not cover \textit{value learning}~\cite{Soares2018ValueLearningProblem}: the problem of eliciting the value alignment of decisions based on human interactions/examples. Value learning implementations remain scarce, though. Exceptions include Anderson et al.~\cite{Anderson2018}, who learn ethical principles via inductive logic programming; Wynn et al.~\cite{learningHumanLikeValues2024representationalalignment}, who propose representational alignment for learning values in LLMs; and some works using deep learning~\cite{wang-etal-2024-interpretableRewardModelingLLM,Peschl2022LearnPreferencesFromExperts,Wang2024MAP:Palette}.

Using the previous terminology, we can classify our work as a method for value learning combined with value system \textit{estimation of/aggregation into} distinct groups in a society. The latter distinguishes our approach from the value inference schema, as value system aggregation has so far considered the aggregation of previously elicited value systems into a single one for the society~\cite{leraleri2024aggregation,AaronAgreementTOPSIS}, while we estimate separate value systems for different groups. 


\subsection{Multi-objective RL for Value Learning}

For value learning~\cite{Soares2018ValueLearningProblem}, a decision-making framework that explicitly represents the alignment of alternatives with multiple values is required. The most widely used is multi-objective decision-making~\cite{Vamplew2018Human-alignedProblem,Smith2023balancinggoalsimportant}. 
In particular, modelling value alignment with utilitarian rewards in (multi-objective) \textit{reinforcement learning}, (MO)RL, is considered a promising approach~\cite{virginiaDignumResponsibleAutonomy2017,Leike2018ScalableAA} and has even been used to approximate other ethical theories such as deontology~\cite{oftenDLIgnacio2025}. 

Value learning in MORL may be tackled via multi-objective inverse reinforcement learning~\cite{kishikawaMOIRL2022,Peschl2022LearnPreferencesFromExperts} (MOIRL), the problem of learning a reward function for each goal/value and estimating the goal-based behaviour of individuals as the policy that maximizes a combination of these rewards, by observing behaviour traces. However, value learning based on these traces is risky, as the learned rewards might only be able to distinguish the \emph{best} value-aligned behaviours from the rest, failing at evaluating suboptimal choices. 

An alternative is \textit{Preference-based Reinforcement Learning} (PbRL)~\cite{surveyRLFromPreferencesPbRL} --or RLHF~\cite{christiano2023deeprlpreferences}--, the problem of learning a RL policy based on trajectory comparisons. 
PbRL often consists of two phases: first, a reward model that captures preferences is learned; second, a policy that maximizes this model is obtained via RL. Thus, these methods are a way to perform inverse RL~\cite{brown2019betterthandemonstratorimitationlearningautomaticallyranked}. A single pass through these phases (\textit{offline} PbRL) may misgeneralize after deployment, except under restrictive conditions~\cite{freehand2024}. As such, \textit{online} PbRL (or human-in-the-loop, HiL) has been suggested, where an agent seeks to maximize positive feedback in interaction with a human using a limited number of queries~\cite{musulmaniSDP2024}. Online feedback is particularly desirable for value learning, as it fosters self-reflection~\cite{Liscio2023Sociotechnical}.


However, single-objective PbRL cannot capture diverse preferences (value systems) or simultaneously represent multiple goals (values). The first limitation has been addressed through personalization and clustering, mainly in the context of generative AI~\cite{Park2024RLHFAggregation,Poddar2024PersonalizingLearning,pmlr-v235-chakraborty24b,zhong2024provablemultipartyreinforcementlearning}. The second has motivated multi-objective PbRL~\cite{Mu2024Preference-basedModeling} (PbMORL), either with manually specified objectives~\cite{Lu2024InferringLearning} or with objectives lacking an explicit AI-usable representation~\cite{wang-etal-2024-interpretableRewardModelingLLM}.

To our knowledge, our approach is the first to address both limitations simultaneously, albeit with trade-offs. Some authors~\cite{Park2024RLHFAggregation,Poddar2024PersonalizingLearning} personalize LLMs via user embeddings; by contrast, our method avoids them, but requires more data for new agents. In~\cite{pmlr-v235-chakraborty24b}, a single policy is learned that aggregates a preference-based clustering of agents, an alternative to our per-cluster policy solution. In~\cite{wang-etal-2024-interpretableRewardModelingLLM}, rewards are learned for several goals and weighed at inference stage to adapt to new users; however, the analysis of user groups is not clarified.  In~\cite{Wang2024MAP:Palette}, the generative model alignment problem is modelled with transparent value constraints, but requires expensive interactions to elicit the feasible/wanted constraints. Finally, in~\cite{Mu2024Preference-basedModeling} , theoretical results on Pareto efficiency in PbMORL are provided, but the estimation of particular user preferences is not contemplated.

\section{Representing Value Systems in MDP}\label{sec:background-valuealignment}
In the following, we adapt and extend our previous model for representing values and value systems in a society~\cite{andresEcai2025} to problems involving sequential decisions, and discuss its properties.

\subsection{Representing Value Alignment}
We set out from a set of $m$ values $V = \{v_1, ..., v_m\}$ , where each value $v_i$ constitutes a label for a human value. When \textit{grounded} in a decision-making domain, a value label acquires a particular 
meaning. Here, we focus specifically on decision-making domains modelled via MDPs. 
In particular, the meaning of each value label is grounded by the notion of the \textit{alignment} of a set of trajectories with this value, defined through a preference relation. Here, a trajectory $\tau$ of length $\abs{\tau}=n$ is a sequence $((s_0,a_0),\dots, (s_{n-1},a_{n-1})) \in (S\times A)^n$ where $S$ is the set of states in the MDP and $A$ the available actions.

\begin{definition}[Value Alignment]\label{def:value-alignment-preference}
     The alignment of a set of trajectories $\mathcal{T}$ with a value $v_i$ (in general, the alignment preferences with $v_i$) is represented by a weak order $\preccurlyeq_{v_i}$ over $\mathcal{T}$, where $\tau\preccurlyeq_{v_i} \tau'$ means that the trajectory  $\tau'$ is at least as aligned with value $v_i$ as $\tau$.
\end{definition}

Inspired by other works in the area~\cite{rodriguez2026reinforcementEthicalEmbedingWeightsRewardRL,Serramia2018,montes2022synthesis}, we use a specific kind of utility function, in our work called \textit{alignment function}, $\A_{v_i}$ to quantify value alignment, i.e. to represent the relation $\preccurlyeq_{v_i}$: i.e., for all $\tau, \tau' \in \mathcal{T} $:
    $\tau \preccurlyeq_{v_i} \tau' \iff \A_{v_i}(\tau) \leq \A_{v_i}(\tau')$. Then, to specify the semantics of a set of values, we define the notion of \textit{grounding}.
\begin{definition}[Grounding]\label{def:grounding}
    A \textbf{grounding} of the set of values $V$ is a set of weak orders $\preccurlyeq_{V} = \left\{\preccurlyeq_{v_i}\right\}_{i=1}^m$. Given the respective alignment functions, a \textbf{grounding function} for $V$ is: $G_V=\left(\A_{v_1}, \dots, \A_{v_m}\right)$. 
\end{definition}


Given that most behaviours in MDP scenarios are modelled through reward functions
, we simplify the set of possible grounding functions to those that can be implemented with a multi-objective reward function vector $\Rv: S\times A \to \R^m$. We say that $\Rv$ implements a grounding function $G_V$ when, for all $\tau\in \T$\footnote{In Eq.~\ref{eq:valuereward}, if there is a general concern to prioritize short-term value alignment, we can consider a cumulative discount factor $\gamma<1$, obtaining $G_V(\tau) = \sum_{i=0}^{|\tau|}  \gamma^i \Rv_{V}(s_i,a_i)$.}:
\begin{equation}\label{eq:valuereward}
G_V(\tau) = \sum_{i=0}^{|\tau|}  \Rv_{V}(s_i,a_i)
\end{equation}

We write $\Rv(s,a)=\left(R_{v_1}(s,a), \hdots, R_{v_m}(s,a) \right)$ to denote the reward vector for a given state-action pair $(s,a)$; and for each $v_i \in V: R_{v_i}:S \times A \to \R$ represents the \textit{value reward}, that is, the alignment of $(s,a)$ with value $v_i$. An MDP with such a reward vector constitutes a Multi-Objective MDP (MOMDP)~\cite{Vamplew2018Human-alignedProblem}.


\subsection{Representing Value Systems}\label{sec:representingvaluesystems}
An agent's \textit{value system} expresses the importance assigned to each value. Given a grounding, it induces preferences over trajectories.

\begin{definition}[Value system]\label{def:value-system}
Let $\preccurlyeq_{V}$ be a grounding for a set of values $V$. The \textbf{value system} of an agent $j$, based on the grounding $\preccurlyeq_{V}$, is determined by a weak order $\preccurlyeq^j_{V}$ over $\mathcal{T}$. If $\tau \preccurlyeq^j_{V} \tau'$, we say that $\tau$ is equally or more aligned than $\tau$ with $j$'s value system.
\end{definition}


Following other work in the field ~\cite{linearscalarizedMOMDP2013,AaronAgreementTOPSIS,leraleri2024aggregation,rodriguez2026reinforcementEthicalEmbedingWeightsRewardRL}, we assume that the value system of an agent can be expressed through a linear combination of the alignment of a trajectory with the values (Definition~\ref{def:value-system-alignment-function}). \highlight{Our framework can then be seen as \textit{welfarist utilitarian}~\cite{Sinnott-Armstrong2019-SINC-5}, where values are conceived as different sources of good, and each agent's value system weighs their relative importance.}


\begin{definition}[Value System Function]\label{def:value-system-alignment-function}
Let $j$ be an agent with value system $\preccurlyeq^j_{V}$ and grounding function $G_V$. The function 
$\A_{W_j,{G_V}}(\tau) = W_j\cdot(\A_{v_1}(\tau), \dots, \A_{v_m}(\tau))^T$ is a \textbf{value system function} for $j$ if it represents $\preccurlyeq^j_{V}$ over $\mathcal{T}$, i.e., for all $\tau,\tau' \in \mathcal{T}$:
$$\A_{W_j,{G_V}}(\tau) \leq \A_{W_j,{G_V}}(\tau') \iff \tau \preccurlyeq^j_{V} \tau'$$ where $W_j=(w_j^{v_1},\dots,w_j^{v_m})$ are the \textit{value system weights} that represent the relative importance of each value. We consider normalized weights in the unit $(m-1)$-simplex: $W_j\in {(0,1)}^m$, $\sum_{i=1}^m w_j^{v_i}=1$.
\end{definition} 
If $G_V$ is implemented by a reward vector, we express $\A_{W_j,{G_V}}$ as:

\begin{equation}
\A_{W_j,G_V}(\tau)=   
\sum_{i=0}^{|\tau|} W_j \cdot \Rv(s_i,a_i)^T.
\label{eq:vsalignasutility}
\end{equation}
Thus, the \textit{value system reward} for each agent becomes a linear scalarization of $\Rv$ with weights $W_j$:
$$R_j(s,a)=W_j\cdot \Rv(s,a)^T$$ 
We also define the value system represented by $R_j$ as $\preccurlyeq^{W_j}_{\Rv}$.

\subsection{Representing the Value System of a Society}

Values are inherently social notions~\cite{schwartz1992universals,Osman2024}, and different agents hold different value systems~\cite{leraleri2024aggregation}. 
We set out from a society of agents $J$, each with its own individual value system $\preccurlyeq_{V}^j$ based on its individual grounding \highlight{$\preccurlyeq_{V,j}=(\preccurlyeq_{v_1,j},\dots,\preccurlyeq_{v_m,j})$.}

Agents can potentially have different views on the meaning of values, \highlight{i.e. their groundings might differ.} However, within
human societies and many application domains, a \textit{social} grounding exists~\cite{andresEcai2025}, i.e. there is a \highlight{near-consensus on the meaning of values. An example in the real world is the medical domain~\cite{userStudyLearningValues}. A consensual grounding is further necessary as a basis for interpretation and comparison of the value systems of agents. Thus, we assume that such a meaningful social grounding is obtainable, i.e. a set of preference relations $\preccurlyeq_{V}=\{\preccurlyeq_{v_1}, \dots,\preccurlyeq_{v_m}\}$ exists that is sufficiently coherent with the variety of individual groundings in the society.} 

\highlight{By contrast, we acknowledge that the importance that each agent gives to each value (their value systems) can vary substantially. This is evidenced, for example, in the world values survey~\cite{WVS_Round7_2022}, where people from different countries hold diverging value-based opinions.} Still, since people in the same society have their value system influenced by culture, we can expect regularities in their value systems across social groups \cite{grenfell2014pierre}. \highlight{This suggests that in our society, finding subgroups of agents that can be represented by a shared (possibly aggregated~\cite{leraleri2024aggregation}) value system is a natural expectation.} 

\highlight{The previous considerations lead us to }define the value system of a society~\cite{andresEcai2025} as the composition of a social grounding with a set of value systems that represent the value-based preferences of different groups of agents in the society. 

\begin{definition}[Value system of a society]\label{def:society-value-system}
\highlight{A \textbf{value system of the society} $J$ (or \textbf{social value system}) is a tuple $(\beta,\Omega,\preccurlyeq_V)$ where $\Omega=\left\{\preccurlyeq^l_{V}\right\}_{l=1}^L$ is a set of $L$ value systems based on a grounding $\preccurlyeq_V$, and $\beta : J \to \{1,\dots, L\}$ is a function that assigns each agent to one value system.
}
We refer to the group of agents assigned, through $\beta$, to the $l$-th value system ($\preccurlyeq^l_{V}$) as the $l$-th \textit{cluster} of the society.

\end{definition}

\section{Social Value System Learning in MDPs}\label{sec:bilevelproblem}\label{sec:proposal}\label{sec:algorithm} 

In this section, we define the problem of learning the value system of a society of agents based on examples of value alignment and value-based preferences, and present an algorithm for solving it. 

\subsection{Social Value System Learning Problem}

We assume that for each agent $j$, we have access to examples of their value system and value alignment preferences over pairs of trajectories. In particular, for each agent $j$ there is a dataset $DS_j$ composed by entries of the type $(\tau, \tau',y^j_V,y^j_{v_1},\dots,y^j_{v_m})$, where $y^j_{\_}\in \{0,0.5,1\}$ indicates as to whether agent $j$ prefers $\tau$ over $\tau'$ (1), $\tau'$ over $\tau$ (0) or is indifferent (0.5), with regard to $j$'s value system ($y^j_V$) and to each of the individual values ($y^j_{v_i}$).
In the sequel, we use $D_j$ to refer to the set of trajectory pairs $\{(\tau, \tau')\}$ that are included in a dataset \highlight{$DS$ and define $DS=\bigcup_{j\in J} DS_j$ and $D=\bigcup_{j\in J}D_j$.}


First , we need to quantify value system and alignment preference differences. To do this, we define the \textit{discordance} between two preference relations by normalizing the number of ordered pairs of trajectories in a set $S\subseteq \T\times \T$ that are ranked differently.
Given two relations $\preccurlyeq^1$ and $\preccurlyeq^2$ \textit{discordance} is calculated as follows:


\begin{equation}\label{eq:discordance}
    d_{S}\left(\preccurlyeq^1, \preccurlyeq^2\right) = \frac{1}{\abs{S}}\sum_{\substack{(\tau,\tau')\in S}}\mathds{1}\left(\left({\tau\preccurlyeq^1 \tau'}\right) \not\equiv \left(\tau \preccurlyeq^2 \tau'\right)\right)
\end{equation}  

       where $\mathds{1}\left(\left({\tau\preccurlyeq^1 \tau'}\right) \not\equiv \left(\tau \preccurlyeq^2 \tau'\right)\right)$ yields 1 when the preference given by $\preccurlyeq^1$ and $\preccurlyeq^2$ over the pair $(\tau,\tau')$ differs, and $0$ otherwise.
       
To quantify the degree by which a candidate social grounding represents the individual groundings of a set of agents (i.e. a given society), we employ the notion of \textit{coherence}.

\begin{definition}[Coherence]\label{def:coherence} Let $\{\preccurlyeq_{v_i,j}|j \in J\}$ be the set of value alignment preferences with value $v_i$ held by each agent in a society $J$. The \textbf{coherence} of an alignment preference $\preccurlyeq_{v_i}$ with regard ${v_i}$ with $\{\preccurlyeq_{v_i,j}|j \in J\}$, over the trajectory pairs $D$ is given by:

\begin{align*}
    \chr_{D}(\preccurlyeq_{v_i}) = 1-\frac{1}{\abs{J}}\sum_{j\in J} d_{D_j}(\preccurlyeq_{v_i}, \preccurlyeq_{v_i,j})
\end{align*}
    
We define the coherence of a grounding  $\Omega = (\preccurlyeq_{v_1}, \dots, \preccurlyeq_{v_m})$ by $\chr_{D}(\preccurlyeq_V) = \frac{1}{m}\sum_{i=1}^m \chr_{D}(\preccurlyeq_{v_i})$.
\end{definition}

\highlight{As we assume that there is a near-consensus in the society concerning value meaning, we assume that a social grounding $\preccurlyeq_{V}$ exists that has a coherence with the agents' alignment preferences that tends to $1$}. In a MDP, this suggests that a reward vector $\pmb{R}$ can be learned to represent $\preccurlyeq_{V}$ with high coherence. \highlight{We denote such learned grounding with $\preccurlyeq_\Rv=(\preccurlyeq_{R_{v_1}},\dots,\preccurlyeq_{R_{v_m}})$}. Additionally, \highlight{we can estimate the social value system using the grounding $\preccurlyeq_{\Rv}$ and a set of $L$ value system weights $\pmb{W} =(W_1,\dots, W_L)$: following Section~\ref{sec:representingvaluesystems}, we estimate the value system of the $l$-th cluster of the society with the value system represented by the reward $R_l(s,a)=W_l\cdot \Rv(s,a)^T$, i.e. for each $l$: $\preccurlyeq_V^l\approx \preccurlyeq^{W_l}_{\Rv}$. Then, each agent's value system $\preccurlyeq_V^j$ is represented with the reward function of its cluster, i.e. $R_j(s,a) \triangleq R_{\beta(j)}(s,a)=W_{\beta(j)}\cdot \Rv(s,a)^T$. We denote the social value system implemented by $\Rv$ and $\pmb{W}$ with $(\beta,\pmb{W},\Rv)$, to refer to the social value system $(\beta,\Omega,\preccurlyeq_V)$ (Definition~\ref{def:society-value-system}) where $\Omega=\{\preccurlyeq_\Rv^{W_l}\}_{l=1}^L$ and $\preccurlyeq_V=\preccurlyeq_\Rv$}.

To evaluate the quality of a value system of the society we use two metrics from our previous work~\cite{andresEcai2025}: \textit{representativeness} (Definition~\ref{def:representativeness}) and \textit{conciseness} (Definition~\ref{def:conciseness}). The first refers to the accuracy with which each agent's value-based preferences are recovered by their assigned value system. The latter measures the differences between the value systems represented in the solution, which motivates the use of fewer clusters to describe the society.

\begin{definition}[Representativeness of a value system of a society]\label{def:representativeness}
    The \textbf{representativeness} of a \highlight{social value system $(\beta,\Omega,\preccurlyeq_V)$ over the trajectory pairs} $D=\bigcup_{j\in J}D_j$ is:


    $$\repr_{D}\left(\beta,\Omega,\preccurlyeq_V\right) = 1-\frac{1}{\abs{J}}\sum_{j\in J}d_{D_j}\left(\preccurlyeq^{\beta(j)}_V, \preccurlyeq_V^j,\right)$$


\end{definition}

\begin{definition}[Conciseness of value system of a society]\label{def:conciseness}
The \textbf{conciseness} of a \highlight{social value system $(\beta,\Omega,\preccurlyeq_V)$ over $D$} is:

 $$\conc_{D}\left(\beta,\Omega,\preccurlyeq_V\right) = \min_{\substack{l\neq l' 
}
} d_{D}(\preccurlyeq_V^l, \preccurlyeq^{l'}_V), $$



\end{definition}

Both metrics are normalized in $[0,1]$ and \highlight{higher values indicate a better representation of the agent's value systems, and higher significance of the clusters composing the social value system (in terms of their ability to represent different preferences), respectively}. 

To tackle the trade-off between conciseness and representativeness, we propose minimizing a heuristic metric $\Gamma(\repr,\conc)$. 
Natural choices for $\Gamma$ are clustering metrics that compare inter-cluster and intra-cluster distances, such as the Dunn-Index~\cite{Dunn01011974} or the Ray-Turi Index~\cite{Ray2000DeterminationSegmentation}
. Here, we propose using a version of the latter, as it can simply be expressed by the ratio $(1-\repr)/\conc$, i.e. the mean of the distances from each agent to the centroids (the value systems in each cluster) over the minimum distance among centroids. As conciseness is likely to reach $0$, and to prioritize representativeness over conciseness, we use $\Gamma(\repr,\conc)=(1-\repr)/(1+\conc)$. The analysis of alternative $\Gamma$ metrics is left for future work. 

We are now in a position to define the \textit{social value system learning} problem, as \highlight{learning a value system of the society $(\beta^*,{\pmb{W}^*},\Rv^*)$} that solves the following bi-level optimization problem:

\begin{align}\label{eq:bilevel}
(\beta^*,\pmb{W}^*) \in \argmin_{\beta,\pmb{W}}& \ \Gamma\left({\repr_{D}\left(\beta,{\pmb{W},\Rv^*}\right)}, {\conc_{D}\left(\beta,\pmb{W},\Rv^*\right)}\right)\nonumber
\\
\text{subject to} \quad \Rv^* &\in \argmax_{\Rv} \ \chr_{D}(\preccurlyeq_{\Rv})
\end{align}
\highlight{Problem~\ref{eq:bilevel} formulates the aim of maximizing the trade-off between representativeness and conciseness, based on finding a grounding with maximal coherence. This ensures that the learned value system weights are based on a correct estimation of the social grounding.}

\highlight{\subsection{Learning Value-aligned Behaviours with HiL}}\label{sec:value-system-learning-in-MDP}



The learned value systems for each cluster should represent well the corresponding agents' behaviours. In MDPs, different behaviours are represented via \textit{policies} that obtain an accumulated reward. 
\highlight{Although human agents may occasionally be misaligned, value-aware software agents should remain consistently value-aligned. Thus, they should follow \textit{rational} value-aligned policies, i.e. those that maximize an accumulated reward that takes into account the grounding of the values.
Assuming that all values are worth being promoted, the (software) agents' policies are, further, expected to be Pareto-efficient regarding this grounding.}


 Given a cluster $l$ described by a reward vector $\Rv$ and weights $W_l$, we can define a policy $\pi_l$ that maximizes  $R_l(s,a)=W_l\cdot \Rv(s,a)$. This policy should be \textit{aligned} with the behaviours of the agents belonging to the cluster $l$, \highlight{and Pareto-efficient regarding the social grounding that $\Rv$ approximates.} 

However, in our setting, we learn the society's value system over a finite dataset of trajectory comparisons of unknown nature. Even if we find cluster value systems with high representativeness, they may not represent well the actual agents' behaviours.
The reason is that the learned rewards do not always \textit{generalize} well \highlight{to the MDP dynamics}. Capturing the agents' choices from generic comparisons might not be enough. Instead, we should try to acquire value systems that not only correctly evaluate suboptimal trajectories, but also tend to prefer the ones that are close to the agents' behaviours.  

To address the last issue, inspired by human-in-the-loop (HiL) PbRL~\cite{musulmaniSDP2024}, we propose \highlight{asking for value-based and alignment preferences over trajectories sampled from estimates of the policies $\pi_l$ while learning a social value system. This ``online'' feedback should steer the learned policies towards the actual agent behaviours}. 

\subsection{Algorithm for Social Value System Learning}



We employ a deep learning solution approach to Problem~\ref{eq:bilevel}. We use two models. First, a reward vector network $\Rv^\theta$ with parameters $\theta$, that represents a social grounding. Second, $L_{max}$ neural networks. Each network consists of a linear layer given by certain value system weights $W_l^\omega$ that are parametrized with \highlight{$\omega \in [\R^{m}]^{L_{max}}$ through a \textit{softmax} calculation: $W^\omega_l=(w_j^{v_1}, \dots, w_j^{v_m}) = \frac{\exp \omega_l}{\sum \exp(\omega_l)} $, so they remain in the simplex. We then write $\pmb{W}^\omega = (W^\omega_{1},\dots,W^\omega_{{L_{max}}})$, but given an assignment $\beta$, we only consider the necessary $L\leq L_{max}$ networks/weights.} 
Then, the value system estimated through these networks would be written as $(\beta,\pmb{W}^\omega,\Rv^\theta)$.

For learning purposes, we represent the grounding preferences as well as the value-based preferences induced by these networks in the datasets using the differentiable Bradley-Terry (BT) model~\cite{bradleyTerryModel1952}. It estimates the relative preference between two trajectories given a reward function (Eq.~\ref{eq:bradley-terry})\footnote{In Eq.~\ref{eq:bradley-terry}, the designer might consider a discount factor $\gamma<1$ as in Eq.~\ref{eq:valuereward}.}. With this model, we consider $p(\tau,\tau'|R) = 0.5$ only if $\tau$ and $\tau'$ are similarly aligned and it tends to $1$ or $0$ if their difference in alignment is increasingly strict. We then use cross-entropy-like loss calculations to reproduce the discordance of our BT model with the dataset preferences. Finally, we use gradient descent in a structured way to update the networks $\pmb{W}^\omega,\Rv^\theta$ in line with the goals in Problem~\ref{eq:bilevel} (see supplementary material for details).

\begin{equation}
    p(\tau,\tau'|R) = \frac{\exp{ \sum_{(s,a)\in\tau} R(s,a)}}{\exp{\sum_{(s,a)\in\tau} R(s,a)}+\exp{ \sum_{(s',a')\in\tau'} R(s',a')}}\label{eq:bradley-terry}
\end{equation}

\highlight{Algorithm~\ref{alg:algorithm-vslp} is our proposed, approximate solution to Problem~\ref{eq:bilevel}. It builds upon }our previous work~\cite{andresEcai2025} (which solves Problem~\ref{eq:bilevel} in non-sequential decision making) 
 and PbMORL~\cite{Mu2024Preference-basedModeling}, a MORL algorithm that estimates a set of Pareto-efficient policies from online preferences. It uses Envelope Q-Learning~\cite{YangAAdaptation} (EQL) to learn a multi-objective policy conditioned on weights: $\Pi(s,a|W)$. This allows to request for the policy $\pi_l$ that maximizes $W_l\cdot \Rv$ with $\pi_l =\Pi(s,a|W_l)$. By sampling different weights we get approximately efficient policies that represent the agents' value systems. Our system introduces key changes, as its goal is not finding the whole front, but rather, a smaller set of weights and policies that represent the agents' values.

 
\highlight{
\textbf{Initialization.} Our algorithm (SVSL-P, Algorithm~\ref{alg:algorithm-vslp}) starts by finding an approximation to Problem~\ref{eq:bilevel} using a static dataset $DS$, to have an initial ``good guess'' of the reward vector and the clusters prior to learning the RL policies. To do so, we execute a version of Algorithm~2 from~\cite{andresEcai2025} (Algorithm~S2 in the supplementary material). The algorithm performs \textit{Expectation Maximization} (see below) with occasional random mutations to avoid local convergence problems~\cite{emlocalconvergence}, and it is run until we reach a certain predefined score $A_{ref}\in[0,1]$ in both grounding coherence and representativeness.}


\highlight{\textbf{Main loop (Lines 3-17)} Now, the aim is to iteratively improve the solution (the social value system $(\beta,\pmb{W}^\omega,\Rv^\theta)$) (\textit{Value system update}) while learning the corresponding MO policy $\Pi(s,a|W)$ (\textit{Policy update}) using HiL with environment experiences (\textit{Exploration}).} 

\highlight{\textbf{Exploration} (Lines 3-5). At every timestep (main loop iteration) we collect a transition $(s,a,\Rv^{\theta}(s,a),s',W)$ and add it to an \textit{experience} replay buffer $R_e$. The action is selected by an $\epsilon$-greedy version of the current policy estimate $\Pi(s,a|W)$ supplying a randomly selected weight combination $W$ from those in $\pmb{W}^{{\omega}}$.}

\highlight{\textbf{Value system update} (Lines 6-12). Every $K$ exploration timesteps, we perform two consecutive tasks:}
    \begin{enumerate}
        \item \highlight{\underline{\emph{Preference collection}} (Lines 7-8). We extract new preference feedback by asking a random group of $N_a$ agents about a set of $N_s$ pairs of trajectories in $R_e$ selected with \textit{QPA}~\cite{QPAhu2024querypolicy}. Namely, we ask each agent for its value alignment and value-based preferences about each trajectory pair, and this data is inserted in a \textit{preference} buffer $R_p$.}   
        \highlight{This step is present in PbMORL, but that algorithm can ask for preferences derived from any weight combination in the whole simplex. The feedback required by our solution is far more simple: an agent is only asked to answer about value alignment and its own value system. }
        \item \highlight{\textit{\underline{Expectation-Maximization (EM)}} (Lines 10-11). Similar to PbMORL, here we improve the reward vector estimate $\Rv^\theta_V$. In our case, we also update the value system weights $\pmb{W}^\omega$ and revise the clustering (updating $L$ and $\beta$), by performing a series of $E_r$ \textit{EM} cycles (see Algorithm~S1 in supplementary material for details). We take $b_{ep}$ random entries from $R_p$ for each agent; add them to the static dataset $DS$; and perform a ``hard'' \textbf{E-step}: the assignment of each agent into the cluster that best represents its value system. Then, we perform the \textbf{M-step} ($m_{r}$ times) to update\footnote{See supplementary material, for details on this update process.} $\Rv^\theta$  and $\pmb{W}^\omega$ based on a random batch of entries of size $b_{mp}$ in 
        $R_p$. After $E_r$ EM cycles, we obtain a new social value system estimation $(\beta,\pmb{W}^\omega,\Rv^\theta)$. We then use $\Rv^\theta$ to update the experience buffer rewards in $R_e$.}
    \end{enumerate}

\highlight{\textbf{Policy update} (Lines 13-16). We apply the EQL Q-learning step from PbMORL so that $\Pi(s,a|W)$ learns to maximize the scalarization of $\Rv^\theta$ under weights and transitions collected from $b_\pi$ experiences in $R_e$, selected via prioritized \textit{hybrid experience replay}~\cite{QPAhu2024querypolicy}\footnote{In particular, we collect $b_\pi/2$ transitions randomly among the most recent ones, and the rest are selected through prioritized experience replay~\cite{schaul2016prioritizedexperiencereplay}.}.}

\highlight{After a number of timesteps $T$, the algorithm returns a social value system $(\beta,\pmb{W}^\omega,\Rv^{\theta})$ and the multi-objective policy $\Pi(s,a|W)$ that models the policies that represent each cluster: $\pi_l=\Pi(s,a|W_l)$. }

\begin{algorithm}[h]
\caption{Social Value System Learning in MDP (\textbf{SVSL-P})}\label{alg:algorithm-vslp}
\RaggedRight
\textbf{Input:} Dataset $DS$. Maximum number of clusters $L_{max}$. Reference accuracy $A_{ref}$. Timesteps between VS updates $K$. Total timesteps $T$. Trajectory sampling size $N_s$. Asked agents per VS update $N_a$.  
Per-agent batch size $b_{ep}$ (for E-steps) and $b_{mp}$ (for M-steps). EQL batch size $b_{\pi}$ and gradient steps $T_\pi$. Parameters of Algorithm~S1 and~S2 in supp. material. 

\textbf{Output:} Assignment of agents into clusters $\beta$, a reward vector $\Rv^\theta$, value systems $\{W^\omega_{l}\}_{l=1}^{L}$ and weight-conditioned policy $\Pi$. 
\begin{algorithmic}[1]
    \State Initialize replay buffers $R_e$, $R_p$ and a Q network $Q(s,a|W)$.
    \State  $\beta_0$, $\Rv^{\theta}$, $\pmb{W}^{\omega}$, $\lambda$ $\gets$ Run \textproc{\textbf{Algorithm~S2}},  supp. material, until $\min_{i}\chr_{D} \geq A_{ref}$, $\repr_{D}\geq A_{ref}$. \footnotemark
    \For{environment timestep $t = 0,\dots, T-1$}
    \State Sample one set of VS weights $w$ from those in $\pmb{W}^\omega$.
    \State Collect transition $(s,a,R^{\theta}_V(s,a),s',d,w)$, add it to $R_e$
    \If{$t \mod K = 0$}
        
        \State $B_a\gets$ Select $N_a$ agents at random (no replacement).
        \State $B_p\gets$ Collect $N_s$ pairs of trajectories from $R_e$ (\textbf{QPA}).
        \State \textit{\underline{Preference collection}}: add new $B_a\times B_p$ entries into $R_p$.
        \State $\beta,\theta,\omega,\lambda'\gets$ \textit{\underline{Expectation-Maximization (EM)}}: \textproc{\textbf{Algorithm~S1}}($\beta, \theta,\omega,\lambda$,$R_p$,$b_{ep}$,$b_{mp}$), supp. material. 
        \State Update $\Rv^{\theta}$ and relabel $R_e$ with it. Update $\pmb{W}^\omega$. 
    \EndIf
    \For{gradient step in $T_\pi$}
    \State Sample a minibatch from replay buffer $R_e$
    \State Apply the \textit{update} step from the EQL paper~\cite{YangAAdaptation} on $Q(s,a|W)$ using a batch of size $b_{\pi}$ from $R_e$ as in~\cite{Mu2024Preference-basedModeling}.
    \EndFor
    \EndFor     
 \State \textbf{return} ( $\beta$, $\pmb{W}^{\omega}$, $\Rv^{\theta}$), $\Pi(s,a|W)$ (from $Q$).
 \end{algorithmic}
\end{algorithm}

\footnotetext{The parameter $\lambda$ referes to the Lagrange multipliers used when updating $\pmb{W}^{\omega}$ and $\Rv^\theta$ to prioritize coherence over representativeness (See supplementary material).}

\section{Evaluation}\label{sec:evaluation}
\newcolumntype{C}{>{\centering\arraybackslash}X}     
\newcolumntype{L}[1]{>{\raggedright\arraybackslash}p{#1}}
\newcolumntype{R}[1]{>{\raggedleft\arraybackslash}p{#1}}
\newcolumntype{Z}[1]{>{\centering\arraybackslash}X}      

We evaluate our proposal in two synthetic MDPs. In each of them, we simulate a society of agents $J$ that considers a social grounding implemented by a ``ground truth'' reward vector $\Rv$, and have a value system determined by specific weights $\{W_j|j\in J\}$. We selected the weights and generated the static dataset $DS$ as follows. We trained EQL~\cite{YangAAdaptation} to convergence in each case (see parameters in the supplementary material). Then, we sampled 50 equally spaced weight combinations in the weight space. We executed the learned policy for each weight ($\Pi(s,a|w)$) in the environment to produce 1000 trajectories. We selected a set of $M_W$ value system weight combinations whose policies form all the points on the Pareto front found by EQL.
We created $M_A$ agents per value system, for a total of $M_A\cdot M_W$ agents, and sampled $200$ trajectories per agent. A proportion of these trajectories ($r_p$) are ``rational'': they are obtained from the $\epsilon$-greedy application of the policy obtained by EQL using the agent's weights. The rest are sampled via a random policy. The selection of $\epsilon$ and $r_p$ varies in environments to get trajectories of varied quality. From these trajectories, per agent, we sampled 200 value alignment (per value) and 200 value-based preference pairs at random. Then, 50\% of the comparisons of each agent are used to form the static dataset $DS$, the rest are left out as a test dataset.

In each environment, we compare the results obtained by two baselines, a state-of-the-art algorithm and our method (\textbf{\textproc{SVSL}-P}).

The first baseline is \textbf{EQL}~\cite{YangAAdaptation} with the original reward vector $\Rv$. We use the execution that generated the agents and datasets. The second baseline is the social value system learning algorithm from \cite{andresEcai2025} (Algorithm~S2 in the supplementary material, \textproc{\textbf{SVSL}}), \highlight{a preference-based reward learning and clustering baseline {(without HiL)}}. To learn the associated policies for the obtained value systems, we run EQL with the learned reward vector with SVSL.

    The state-of-the-art related algorithm that we analyse is \textbf{PbMORL}~\cite{Mu2024Preference-basedModeling}. It learns a reward vector and a weight-dependent policy $\Pi(s,a|W)$. Since the algorithm does not extract clusters, we first select $\abs{J}$ equally spaced weight combinations in the simplex to represent potential cluster value systems. Then, we use the E-step from Algorithm~S1 to assign each agent to the  cluster that best represents its value system and discard the empty ones. 

We run the algorithms 10 times (different seeds) with the same datasets. The maximum number of clusters for \textproc{SVSL} and \textproc{SVSL-P} was set to $L_{max}=10$ and $L_{max}=15$ for the FF and the MVC environments, respectively. The hyperparameter specifications for all algorithms are detailed and analysed in the supplementary material.

\subsection{Firefighters Environment}
In the firefighters environment (FF), adapted from~\cite{osman2025instillingorganisationalvaluesfirefighters} and tested in previous works~\cite{andres2026JAAMASfinal}, different agents (firefighters) are trained to rescue people and put out a fire in a building, remaining aligned with two values: \textit{professionalism} ($p_f$) and \textit{proximity} ($p_x$). The first promotes behaving under firefighters' best practices, while the latter promotes actions with the goal of saving lives at all costs. The environment consists of $5$ actions and $400$ possible states with ground-truth value rewards bounded in $[-1,1]$. The environment specification details are available in the supplementary material. 

The reward vector model observes a one hot-encoded version of the state-action features (in supplementary material). We use a neural network for each value. Each one consists of three fully-connected hidden linear layers (128 neurons each) followed by an output layer (one neuron, no bias) and $Tanh$ activation functions. We use the same network configuration for all algorithms. 

Using EQL we obtained $M_W=5$ value systems whose policies form the convex part of the Pareto front. We simulated $M_A=3$ agents per value system and sampled 200 trajectories with each policy. The proportion of rational trajectories was set to $r_p=80\%$, and those were sampled using $\epsilon$-greedy policies with $\epsilon=0.1$.

\begin{figure}[h]
    \centering
    \includegraphics[width=0.815\linewidth,trim=0.25cm 0.35cm 0.3cm 1.05cm, clip]{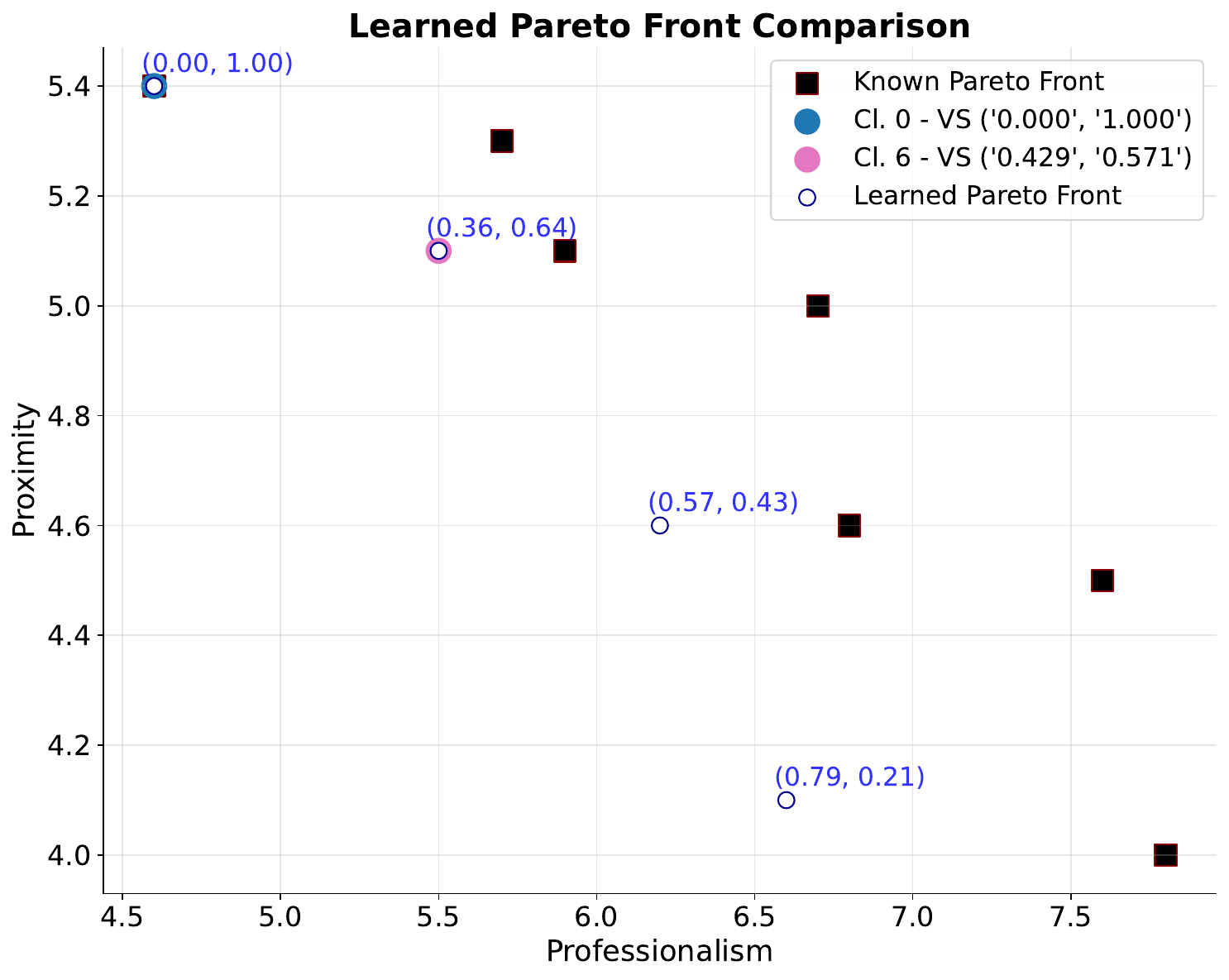}
    \includegraphics[width=0.815\linewidth,trim=0.25cm 0.35cm 0.3cm 1.05cm, clip]{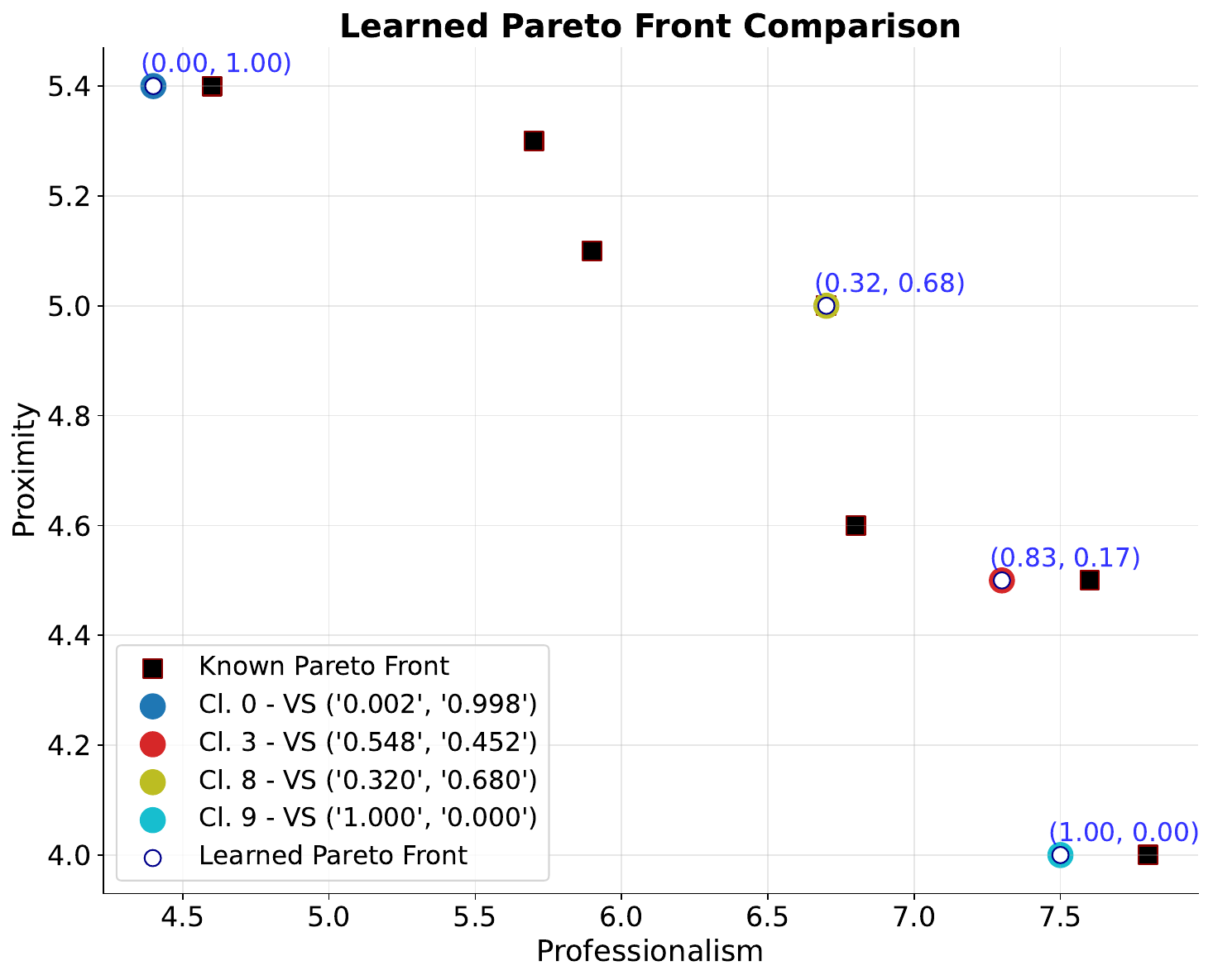}
    \caption{FF environment. Approximated Pareto front and clusters learned with PbMORL (Top) and SVSL-P (bottom, ours) with a particular seed. Black squares form the ground-truth Pareto front. White dots depict weights which policies are in the approximated front. Coloured dots indicate the policies representing each learned cluster (in the legend).}
    \label{fig:paretoeqlff}
\end{figure}

\begin{table*}[h]
\centering
\begin{tabularx}{\textwidth}{L{1.15cm}%
L{1.15cm}%
C
C
C
C
C
|C
C
C
C
C
C
}
\toprule
Method (in FF) & L (histogram) & $\repr$ $\uparrow$ & Prof. $\chr$ $\uparrow$ & Prox. $\chr$ $\uparrow$ & $\conc$ $\uparrow$ & Ray-Turi $\downarrow$ & PF size (all) $\sim$  & PF size (cls.) $ \sim$ & HV. (all) $ \uparrow$ & HV    (cls.) $ \uparrow$ & MUL $\downarrow$ (all) & MUL $\downarrow$ (cls.) \\
\midrule
EQL & - & - & - & - & -  & - & 5 & 5 & 40.52 & 40.52 & 0.0 
& 0.0 \\
SVSL     & \raisebox{-0.4\height}{%
  \makebox[\linewidth]{%
    \begin{tikzpicture}[scale=0.4, baseline=(current bounding box.south)]
      \fill[blue!50] (0,0) rectangle (0.7,1.1);   

      \node[scale=0.6] at (0.35,1.35) {100\%};

      \node[scale=0.8, inner sep=0.2pt] at (0.35,-0.2) {2};

      \path[use as bounding box] (0,-0.25) rectangle (0.7,1.4);
    \end{tikzpicture}%
  }%
}%
         & \makecell{0.915\\ ±0.022}    & \makecell{0.860 \\ ±0.042}      & \makecell{0.858 \\ ±0.019} & \makecell{\textbf{0.163}\\ ±0.055} & \makecell{0.074\\ ±0.024} & \makecell{1.1 \\ ±0.3} & \makecell{2.0 \\ ±0.0} & \makecell{0.0 \\ ±0.0} & \makecell{0.0 \\ ±0.0} & \makecell{45.118 \\ ±0.95}  & \makecell{46.653 \\ ±1.314}  \\
PbMORL  & \raisebox{-0.4\height}{%
  \makebox[\linewidth]{%
    \begin{tikzpicture}[scale=0.4, baseline=(current bounding box.south)]
      \fill[blue!50] (0,0) rectangle (0.7,1.2);    
      \fill[blue!50] (0.9,0) rectangle (1.6,0.72); 
      \fill[blue!50] (1.8,0) rectangle (2.5,0.48); 

      \node[scale=0.6] at (0.35,1.35) {50\%};
      \node[scale=0.6] at (1.25,0.90) {30\%};
      \node[scale=0.6] at (2.15,0.66) {20\%};

      \node[scale=0.8, inner sep=0.2pt] at (0.35,-0.2) {2};
      \node[scale=0.8, inner sep=0.2pt] at (1.25,-0.2) {3};
      \node[scale=0.8, inner sep=0.2pt] at (2.15,-0.2) {4};

      \path[use as bounding box] (0,-0.25) rectangle (3.0,1.35);
    \end{tikzpicture}%
  }%
}%
& \makecell{0.922 \\ ±0.008} 
& \makecell{0.863 \\ ±0.009} 
& \makecell{0.920 \\ ±0.012} 
& \makecell{0.067 \\ ±0.047} 
& \makecell{0.074 \\ ±0.010} 
& \makecell{3.8 \\ ±0.600} 
& \makecell{{2.4} \\ ±0.489} 
& \makecell{35.35 \\ ±0.934} 
& \makecell{31.39 \\ ±2.98} 
& \makecell{0.670 \\ ±0.154} 
& \makecell{1.240 \\ ±0.579} \\

SVSL-P   & \raisebox{-0.4\height}{%
  \makebox[\linewidth]{%
    \begin{tikzpicture}[scale=0.4, baseline=(current bounding box.south)]
      \fill[blue!50] (0,0) rectangle (0.7,1.2);   
      \fill[blue!50] (0.9,0) rectangle (1.6,0.3); 

      \node[scale=0.6] at (0.35,1.35) {80\%};
      \node[scale=0.6] at (1.25,0.45) {20\%};

      \node[scale=0.8, inner sep=0.2pt] at (0.35,-0.2) {4};
      \node[scale=0.8, inner sep=0.2pt] at (1.25,-0.2) {5};

      \path[use as bounding box] (0,-0.25) rectangle (2.2,1.6);
    \end{tikzpicture}%
  }%
}%

         & \makecell{\textbf{0.968} \\ ±0.010} 
& \makecell{\textbf{0.967} \\ ±0.014} 
& \makecell{\textbf{0.951} \\ ±0.009} 
& \makecell{0.049 \\ ±0.016} 
& \makecell{\textbf{0.03} \\ ±0.010} 
& \makecell{3.6 \\ ±0.489} 
& \makecell{3.8 \\ ±0.400} 
& \makecell{\textbf{38.98}\\ ±0.170} 
& \makecell{\textbf{38.83}\\ ±1.454} 
& \makecell{\textbf{0.117}\\ ±0.150} 
& \makecell{\textbf{0.174}\\ ±0.293}\\
\bottomrule
\end{tabularx}

\begin{tabularx}{\textwidth}{L{1.15cm}%
L{1.25cm}%
C
C
C
C
C
C
|C
C
C
C
C
C
C
}
\toprule
Method (in MVC) & L (histogram) & $\repr$ $\uparrow$ & Ach. $\chr$ $\uparrow$ & Safe. $\chr$ $\uparrow$ & Comf. $\chr$ $\uparrow$ & $\conc$ $\uparrow$ & Ray-Turi $\downarrow$  & PF size (all) $\sim$  & PF size (cls.) $ \sim$ & HV (all) $\uparrow$ & HV (cls.) $\uparrow$ & MUL $\downarrow$ (all) & MUL $\downarrow$ (cls.)  \\
\midrule

EQL & - & - & - & - & - & - & - &  14 & 14  & 1.233 & 1.233 & 0.0 & 0.0 \\

SVSL     & \raisebox{-0.4\height}{%
  \makebox[\linewidth]{%
    \begin{tikzpicture}[scale=0.4, baseline=(current bounding box.south)]
      \fill[blue!50] (0,0) rectangle (0.5,1.2);   
      \fill[blue!50] (0.7,0) rectangle (1.2,0.72); 
      \fill[blue!50] (1.4,0) rectangle (1.9,0.24); 
      \fill[blue!50] (2.1,0) rectangle (2.6,0.24); 

      \node[scale=0.6] at (0.25,1.35) {50\%};
      \node[scale=0.6] at (0.95,0.87) {30\%};
      \node[scale=0.6] at (1.65,0.39) {10\%};
      \node[scale=0.6] at (2.35,0.39) {10\%};

      \node[scale=0.8, inner sep=0.2pt] at (0.25,-0.2) {1};
      \node[scale=0.8, inner sep=0.2pt] at (0.95,-0.2) {2};
      \node[scale=0.8, inner sep=0.2pt] at (1.65,-0.2) {4};
      \node[scale=0.8, inner sep=0.2pt] at (2.35,-0.2) {5};

      \path[use as bounding box] (0,-0.25) rectangle (2.6,1.4);
    \end{tikzpicture}%
  }%
}%

         & \makecell{0.771 \\ ±0.112}       & \makecell{0.775 \\ ±0.134}      & \makecell{0.682  \\ ±0.139} & \makecell{0.628 \\ ±0.165} & \makecell{\textbf{0.051}\\ ±0.031} & \makecell{0.139\\ ±0.053} & \makecell{4.4 \\ ±1.43} & \makecell{2.0 \\ ±0.0} & \makecell{0.556 \\ ±0.510} & \makecell{0.475 \\ ±0.462} & \makecell{14.763 \\ ±11.24}  &
         \makecell{18.562 \\ ±12.06}\\

PbMORL  & \raisebox{-0.4\height}{%
  \makebox[\linewidth]{%
    \begin{tikzpicture}[scale=0.4, baseline=(current bounding box.south)]
      \fill[blue!50] (0,0) rectangle (0.5,0.3);   
      \fill[blue!50] (0.7,0) rectangle (1.2,0.6); 
      \fill[blue!50] (1.4,0) rectangle (1.9,0.6); 
      \fill[blue!50] (2.1,0) rectangle (2.8,1.2); 
      \fill[blue!50] (3.0,0) rectangle (3.5,0.3); 
      
\node[scale=0.6] at (0.25,0.45) {10\%};
      \node[scale=0.6] at (0.95,0.75) {20\%};
      \node[scale=0.6] at (1.65,0.75) {20\%};
      \node[scale=0.6] at (2.45,1.35) {40\%};
      \node[scale=0.6] at (3.25,0.45) {10\%};

      \node[scale=0.7, inner sep=0.2pt] at (0.25,-0.15) {5};
      \node[scale=0.7, inner sep=0.2pt] at (0.95,-0.15) {6};
      \node[scale=0.7, inner sep=0.2pt] at (1.65,-0.15) {7};
      \node[scale=0.7, inner sep=0.2pt] at (2.45,-0.15) {8};
      \node[scale=0.7, inner sep=0.2pt] at (3.25,-0.15) {9};

      \path[use as bounding box] (0,-0.2) rectangle (3.0,1.5);
    \end{tikzpicture}%
  }%
}
& \makecell{0.890 \\ ±0.013} 
& \makecell{0.884 \\ ±0.038} 
& \makecell{0.750 \\ ±0.044} 
& \makecell{0.649 \\ ±0.043} 
& \makecell{0.031 \\ ±0.006} 
& \makecell{0.130 \\ ±0.014} 
& \makecell{15.6 \\ ±2.8} 
& \makecell{7.2 \\ ±1.167} 
& \makecell{\textbf{1.217} \\ ±0.011} 
& \makecell{\textbf{1.191} \\ ±0.024} 
& \makecell{\textbf{1.102} \\ ±0.387} 
& \makecell{1.784 \\ ±0.668}\\

SVSL-P   & 
\raisebox{-0.4\height}{%
  \makebox[\linewidth]{%
    \begin{tikzpicture}[scale=0.4, baseline=(current bounding box.south)]
      \fill[blue!50] (0,0) rectangle (0.7,0.9);   
      \fill[blue!50] (0.9,0) rectangle (1.6,1.2); 
      \fill[blue!50] (1.8,0) rectangle (2.5,0.9); 

      \node[scale=0.6] at (0.35,1.05) {30\%};
      \node[scale=0.6] at (1.25,1.35) {40\%};
      \node[scale=0.6] at (2.15,1.05) {30\%};

      \node[scale=0.8, inner sep=0.2pt] at (0.35,-0.2) {4};
      \node[scale=0.8, inner sep=0.2pt] at (1.25,-0.2) {5};
      \node[scale=0.8, inner sep=0.2pt] at (2.15,-0.2) {6};

      \path[use as bounding box] (0,-0.4) rectangle (2.0,1.2);
    \end{tikzpicture}%
  }%
}%

         & \makecell{\textbf{0.908}\\ ±0.013} & \makecell{\textbf{0.886}\\ ±0.015} & \makecell{\textbf{0.883}\\ ±0.009} & \makecell{\textbf{0.748 }\\ ±0.023 }& \makecell{0.044\\ ±0.012} & \makecell{\textbf{0.088}\\ ±0.013} & 
         \makecell{8.9\\ ±1.375 }
         & \makecell{5.0\\ ±0.775} &
         \makecell{ 1.197\\ ±0.017 } & \makecell{ 1.175 \\ ±0.026}   & 
          \makecell{1.152\\ ±0.542} & \makecell{\textbf{1.522}\\ ±0.644}
         \\
\bottomrule
\end{tabularx}
\caption{FF (top) MVC (bottom). For 10 random seeds, number of clusters obtained and frequency ($L$), representativeness ($\repr$), coherence for each value ($\chr$), conciseness ($\conc$), and Ray-Turi index running each algorithm with 10 random seeds over the test-sets. In the right part of the charts, cardinality (PF size), hypervolume (HV) and Maximum Utility Loss (MUL) of the obtained Pareto fronts using all candidate value system weights (``all'' variant); and of the Pareto front obtained using only the learned weights for the value system of the societies (``cls.'' variant). Results indicate the average values and standard deviations.}
\label{tab:acc:comparison}
\end{table*}

\subsection{Multivalued Car Environment}
The Multivalued Car Environment (MVC) was proposed in~\cite{rodriguez2026reinforcementEthicalEmbedingWeightsRewardRL} to illustrate the problem of learning norm-abiding and value-aligned policies given a computational value specification. In MVC, a car agent wants to reach a destination (value of \textit{achievement}) in a road grid while promoting the value of \textit{safety} by respecting pedestrians and avoiding bumpy areas for the sake
of \textit{comfort}. 

We use the same neural network architecture per value as in FF to estimate the reward vectors. The final layer of each network is left without an activation function (thus, also, unbounded). To avoid overflows, we clamp the learned rewards in the interval $[-100,100]$. Also, due to the fact that the environment episodes are of very different lengths, we introduced a discount factor $\gamma=0.99$ to calculate the alignment of trajectories with values/value systems.

We obtained $M_W=14$ value systems and Pareto-optimal policies with EQL. We created $M_A=2$ agents per value system in the front. To build $DS$, we sampled 200 trajectories with $r_p=80\%$, and $\epsilon=0.1$.

\subsection{Discussion of results}

We first discuss the value systems of the society learned by \textproc{SVSL}, PbMORL and \textproc{SVSL}-P in terms of number of clusters, representativeness, coherence and conciseness over the test set. We focus on the left part of the charts in Table~\ref{tab:acc:comparison} (FF at the top, MVC at the bottom). 

SVSL obtains relatively low representativeness (accuracy in representing the individual value systems of agents) and grounding coherence (accuracy of the learned grounding to represent the simulated one) in the test sets, which implies that it did not generalize to the test dataset. PbMORL and SVSL-P generalized better in terms of representativeness, at the expense of conciseness. SVSL-P in particular, outperforms PbMORL in terms of grounding coherence, as it focuses explicitly on this aspect of the optimization. 
\highlight{SVSL achieves, though, the highest conciseness (relevancy of the cluster value systems learned), but tends, wrongly, to obtain less clusters than required.} 
Notably, in the FF case, SVSL-P achieves a higher representativeness with a similar conciseness than PbMORL while using more clusters, indicating that SVSL-P better recognises the value system diversity. In the MVC domain, though, PbMORL selects significantly more clusters than SVSL-P 
, but this does not result in better representativeness for PbMORL. The Ray-Turi index quantifies the trade-off between representativeness and conciseness: it is smaller in SVSL-P, confirming its best performance.

We now test whether the learned reward vectors induce policies that are aligned with the simulated agents' value systems. Since the simulated agents follow the policies that form the convex part of the Pareto front with respect the ``ground-truth'' grounding, we assess how well the learned policies approximate this front (even if using clusters). The right part of the charts in Table~\ref{tab:acc:comparison}, shows quantitative results on the approximated Pareto fronts across methods and environments. We analyse two fronts. The first one is composed by the policies in $\Pi(s,a|W)$ obtained with the weights that each algorithm considers as potential clusters. For SVSL-P and SVSL, these are the value system weights $\pmb{W}^\omega$, and for PbMORL, the set of equally spaced weight combinations of size $\abs{J}$ (15 in FF, 28 in MVC). The second front is composed only by the policies learned for the weights finally selected as society clusters. The metrics for the first front are labelled as (``all''), and for the second as (``cls.''). 

We use three metrics in Table~\ref{tab:acc:comparison} to describe the fronts: \textit{cardinality} (PF size), \textit{hypervolume} (HV)  and \textit{maximum utility loss} (MUL)~\cite{paretoconditionednetworksMUL}. HV is computed from reference points $(0, 0)$ in FF and $(-40.0, -50.0, -50.0)$ in MVC (HV is scaled down by $10^{5}$ in MVC). MUL is measured against the ground-truth front in FF, and against the front learned with EQL in MVC, as the true front is unknown. Lower MUL indicates closer convergence to these fronts. Higher HV values capture both better convergence and distribution of the cluster policies across the learned front. 

SVSL produces incompetent policies in both MDPs, as its learned reward vectors failed to generalize to their dynamics. In contrast, PbMORL and SVSL-P achieve HV and MUL comparable to EQL when considering the full fronts, impressively, despite the bigger size of the PbMORL fronts. Notably, considering the front obtained by the learned clusters alone, the performance in terms of HV of both algorithms is similar, and SVSL-P achieved a better MUL. This occurs even in MVC, where the front of SVSL-P has a smaller size. In FF, though, its size is bigger: but this does not indicate a bad clustering performance, since HV and MUL metrics outperform PbMORL's, showing SVSL-P clusters are of higher significance. 

Figure~\ref{fig:paretoeqlff} represents the Pareto fronts of PbMORL and SVSL-P estimated from their respective potential value system weights in the FF domain, as well as the evaluation of the policies obtained for each used cluster for a certain seed\footnote{We omitted the fronts in MVC given their 3D visualization complexity. The corresponding graph for every seed is available in the supplementary material.}. Both algorithms approximate well the behaviours in the original front. In both cases, the learned clusters tend to be Pareto optimal for both algorithms, but their distribution across the front is worse for PbMORL. PbMORL typically used only two clusters (Table~\ref{tab:acc:comparison})  to represent the agents' value-based preferences, which means the reward vector was not learned in a sufficiently detailed manner to differentiate them properly.

There are also limitations to analyse. First, there are cases where SVSL-P learns two value systems that represent distinct preferences, yet they induce the same policy (supplementary material, Figures~2,3). Second, the policies learned were generally not exactly in the ground-truth front (obtained by EQL), despite being ``close'' (shown by the MUL metric). In the FF domain, for example, the proportion of cluster weights whose policies were in the ground-truth front was, on average, 47.1\% for SVSL-P and 40.8\% for PbMORL. Lastly, the standard deviations in the front-related metrics (Table~\ref{tab:acc:comparison}, right) are high for PbMORL and SVSL-P. This implies that there is some instability across seeds. However, the number of learned clusters is more stable with SVSL-P. Further analysis on these limitations is available in the supplementary material.

\section{Conclusions}\label{sec:conclusions}

We proposed a computational model of the value systems of a society of agents in the context of Markov Decision Processes (MDP). Given a set of human values, we grounded value alignment in a particular MDP with a multi-objective reward vector and represented the value systems of different subgroups of agents (clusters) via linearly scalarized reward functions. \highlight{We put forward an algorithm that learns an instance of this model from online pairwise trajectory comparisons, that are provided by each agent based on both i) its understanding of value alignment (with each value) and ii) the agent's value system. In parallel, it learns an approximately Pareto-efficient MDP policy for each cluster (in terms of value alignment) that represents behaviours aligned with the value systems of its members.} 
Our algorithm, SVSL-P, is based on a Preference-based MORL (PbMORL) method~\cite{Mu2024Preference-basedModeling} and a previous clustering approach~\cite{andresEcai2025}.

\highlight{The results in two synthetic environments show that SVSL-P can learn a concise, representative and coherent set of value systems to describe a society, and the associated value-aligned behaviours.} The learned models generalize in the environment, leading to approximately Pareto efficient policies in terms of the simulated value alignment specifications. The Pareto fronts are focused on the value systems of the agents, and are equally competent to the fronts derived by PbMORL. Our algorithm also requires less intensive human feedback: SVSL-P only asks each agent about their own value systems and their understanding of values. This last advantage is crucial to the applicability of this type of research in real use cases.

There are limitations and avenues for future work. First, the variability across runs should be reduced. Second, the learned policies are not always Pareto efficient regarding value alignment for all preference-based clusters. 
Third, reducing the number of online queries is needed for real-world applications. For future work, we will revise our methodology to control the obtention of more concise versus representative solutions. \highlight{Although we have tested this framework with real-world data in non-sequential decision making~\cite{andresEcai2025}}, to broaden the applicability to more \highlight{realistic} scenarios, we finally suggest researching on modelling non-linear \highlight{and context-dependent value systems (e.g. by selecting distinct value systems under identifiable conditions)} and agent-based groundings/rewards.




\begin{acks}

This work has been supported by grant COSASS: PID2021-123673OB-C32 funded by MCIN/AEI/10.13039/501100011033 and by “ERDF A way of making Europe”, and by project grant EVASAI: PID2024-158227NB-C32 funded by MICIU/AEI/10.13039/501100011033/FEDER, UE. Andr{\'e}s Holgado-S{\'a}nchez has received funding by grant ``Contratos Predoctorales de Personal Investigador en Formaci{\'o}n en Departamentos de la Universidad Rey Juan Carlos (C1 PREDOC 2025)'', funded by Universidad Rey Juan Carlos.

\end{acks}



\bibliographystyle{ACM-Reference-Format}
\bibliography{aamas2026}

\clearpage

\setcounter{algorithm}{0}
\renewcommand\thealgorithm{S\arabic{algorithm}}
\section*{Supplementary material}

\subsection*{Source Code}
A static version is in the OpenReview submission~\cite{holgado-sanchez2026learning}, and is also available on GitHub\footnote{\url{https://github.com/andresh26-uam/ValueLearningInMOMDP/tree/AAMAS2026}}.

\subsection*{Experimental Details}

In Table~\ref{tab:hyperparameters}, we include a comprehensive table of hyperparameters used in our experiments. See also the glossary tables explaining each hyperparameter, in Tables~\ref{tab:hp-glossary-EQL},~\ref{tab:hp-glossary-SVSL} and~\ref{tab:hp-glossary-SVSLP-PbMORL}. We ran the experiments with 10 seeds.

Importantly, the number of timesteps for PbMORL was always set significantly larger than in SVSL-P to compensate the fact that SVSL-P has an initial process to initialize the reward vector with the static datasets, that is non-existent in PbMORL.

\begin{table*}[t]
    \centering
    \resizebox{\textwidth}{!}{\begin{tabular}{c|rrrr|rrrr}
    \toprule
    Env|Alg & FF | EQL & FF | PbMORL & FF | SVSL & FF | SVSL-P & MVC | EQL & MVC | PbMORL & MVC | SVSL & MVC | SVSL-P \\
    
    \midrule 
    $L_{max}$ & - & 10 & 15 & 10 & - & 28 (number of agents) & 15 & 15  \\  
    $\text{\textit{mrt}}$ & - & - & 0.25 & 0.25 & - & - & 0.25 & 0.25  \\    
    $\lambda$ & - & - & 1.0 & 1.0 & - & - & 1.0 & 1.0 \\
    $\alpha_\lambda$ & - & - & 0.05 & 0.05 & - & - & 0.05 & 0.05 \\
    $\gamma_\lambda$ & - & - & $5\times 10^{-5}$  & $5\times 10^{-5}$ & - & - & $5\times 10^{-5}$  & $5\times 10^{-5}$ \\
    $\alpha_\theta$ & - & 0.0003 & 0.0003 & 0.0003 & - & 0.0003 & 0.0003 & 0.0003 \\
    $\alpha_\omega$ & - & 0.005 &  0.005 & 0.005  & - & 0.005 &  0.005 & 0.005 \\
    
    $I$ / $A_{ref}$ & - & - & 100 & $0.85$ & - & - & 100 & $0.85$ \\
    $N$ & - & - & 5 & 5  & - & - & 8 & 8 \\
    $E_r$ & - & - & 2 & 2 & - & - & 2 & 2 \\
    $m_r$ (gradient steps) & - & 3 & 3 & 3 & - & 3 & 3 & 3 \\
    $p_m$ & - & - & 0.1 & 0.1 & - & - & 0.1 & 0.1  \\
    $s_m$& - & - & 0.1 & 0.1 & - & - & 0.1 & 0.1 \\
    
    $T_i$ & - & 10000 & - & - & - & 10000 & - & - \\
    $b$ & - & 256 & - & - & - & 256 & - & -\\
    $K$ & - & 500 & - & 500 & - & 800 & - & 800 \\
    $N_s$ & - & 300 & - & 300 & - & 200 & - & 200 \\
    $N_w$ ($N_a$ in SVSL-P) & 5 & 10 & - & 11 & 10 & 15 & - & 15 \\
    $b_{ep}$ & - & - & - & 50 & - & - & - & 40\\
    $b_{mp}$ & - & - & - & 50 & - & - & - & 40\\
    $S_p$ & - & 100000 & - & 10000 & - & 80000 & - & 80000\\
    $\gamma$ & 1.0 & 1.0 & 1.0 & 1.0 & 0.99 & 0.99 & 0.99 & 0.99 \\
    $\alpha_{EQL}$ & 0.0005 & 0.0007 & 0.0007 & 0.0007 & 0.0005 & 0.0007 & 0.0007 & 0.0007 \\
    $T$ & 120000 & 250000 & 200000 & 200000 & 150000 & 250000 & 200000 & 200000 \\
    $h_0,h_\infty$ & 0.0,1.0 & 0.05,0.9 & 0.05,0.5 & 0.05,0.9 & 0.0,1.0 & 0.0,0.9 & 0.05,0.95 & 0.05,0.9 \\
    $\epsilon_0,\epsilon_\infty$ & 0.5,0.0 & 0.5,0.05 & 0.5,0.05 & 0.5,0.05 &  1.0,0.0 & 0.8,0.05 & 1.0,0.05 & 0.8,0.05  \\
    $T_\pi$ & 1 & 2 & 2 & 2 & 1 & 2 & 2 & 2 \\
    $\tau$ (or $t_u$) & 1500 ($t_u$) & 0.0001 & 0.0001 & 0.0001 & 1000 ($t_u$) & 0.0001 & 0.0001 & 0.0001 \\
    $b_\pi$ & 32  & 256 & 256 & 256 & 128  & 128 & 128 & 128 \\
    $S_e$ & 100000  & 256000 & 500000 & 500000 & 200000  & 500000 & 500000 & 500000 \\
    $\alpha_{per}$ & no  & 0.6 & 0.6  & 0.6  & no  & 0.6  & 0.6  & 0.6  \\
    $\epsilon_{per}$ & no  & 0.01 & 0.01  & 0.01  & no  & 0.01  & 0.01  & 0.01  \\
    $U_w$ & false  & true  & false & true  & false  & true  & false & true \\
    \bottomrule
    \end{tabular}}
    \caption{Hyperparameters used in each environment per algorithm.}
    \label{tab:hyperparameters}
\end{table*}

\begin{table}[h]
    \centering
    \begin{tabular}{lp{6.5cm}}
        \toprule
        \textbf{Symbol} & \textbf{Description} \\
        \midrule
        $\alpha_{EQL}$ & Learning rate for the $Q$-net.\\
        $T$ & Total number of environment timesteps (and training timesteps).\\  
        $h_0,h_\infty$ & Initial and final homotopy weight (linearly increased per timestep) \\
        $\epsilon_o,\epsilon_\infty$ & Initial and final $\epsilon$-greedy parameter (linearly decayed per timestep) \\
        $T_\pi$ & Number of policy learning gradient steps\\
        $\tau$ & Polyak target network update rate (small, from $0$ to $1$)\\
        $t_u$ & If Polyak is set to $1$, this is used instead: as the number of timesteps to fully override the current target network with the learned one.\\
        $b_{\pi}$ & Batch size of experiences $(s,a,s',r,d)$ from $R_e$ for the $Q$-net update steps.\\
        $S_e$ & Maximum size of the experience replay buffer size $R_e$.\\
        $N_w$ & Number of sampled weights for envelope update (coincides with the homonym parameter in PbMORL and SVSL-P) \\
        $\alpha_{per}$ & Prioritized experience replay (PER) priority (default 0.6).\\
        $\epsilon_{per}$ & Minimum value for priority in PER. Set to 0.01.\\
        $U_w$ & Whether to save sampled weights in $R_e$ for use in update steps instead of sampling new random ones. It is set to ``true'' in SVSL-P and PbMORL, but we found it works better as false for vanilla Envelope Q Learning with static reward functions. \\ 
        \bottomrule
    \end{tabular}
    \caption{Glossary of hyperparameters of Envelope Q-Learning~\cite{YangAAdaptation}. Most are used in SVSL-P and PbMORL.}
    \label{tab:hp-glossary-EQL}
\end{table}
\begin{table}[h]
    \centering
    \begin{tabular}{lp{6.5cm}}
        \toprule
        \textbf{Symbol} & \textbf{Description} \\
        \midrule
        $L_{\max}$ & Maximum number of clusters/components \\
        $\text{\textit{mrt}}$ & Initial mutation probability (probability of mutating a solution) \\
        $\lambda$ & Initial Lagrange multipliers \\
        $\alpha_\lambda$ & Learning rate for Lagrange multipliers \\
        $\gamma_\lambda$ & Decay rate for Lagrange multipliers \\
        $\alpha_\theta$ & Learning rate for the reward model $\pmb{R}^\theta$ parameters $\theta$ \\
        $\alpha_\omega$ & Learning rate for the value system weights $W_l^{\omega}$ parameters $\omega$ \\
        $I$ & Number of reward training iterations in Algorithm~\ref{alg:algorithm2} (or use $A_{ref}$) instead)\\
        $N$ & Size of the clustering candidate list/memory\\
        $E_r$ & Number of times to run the EM-algorithm at each iteration \\
        $m_r$ & Number of M-Step repetitions (after each E-Step) \\
        $p_m$ & Agent reassignment probability (of moving an agent to another cluster) \\
        $s_m$ & Mutation scale for network parameters\\
        \bottomrule
    \end{tabular}
    \caption{Glossary of hyperparameters used in the experiments of SVSL (Algorithm~\ref{alg:algorithm2}). Some are used by SVSL-P.}
    \label{tab:hp-glossary-SVSL}
\end{table}

\begin{table}[h]
    \centering
    \begin{tabular}{lp{6.5cm}}
        \toprule
        \textbf{Symbol} & \textbf{Description} \\
        \midrule
        $T_i$ & In PbMORL: initial amount of random action sampling steps in the environment to fill the replay buffer before any $Q$-net update.\\
        $b$ & Batch size of preferences for the reward learning step in PbMORL (bulk, not per agent or per weight).\\
        $K$ & Number of training timesteps in PbMORL and SVSL-P until calling the agents for new preferences.\\
        $N_s$ & Number of pairs of trajectories sampled every $K$ timesteps to be labelled by users both in PbMORL and SVSL-P.\\
        $N_w$ & Number of random weights to ask the overseer for preferences over the previous $N_s$ trajectories, in PbMORL.\\
        $N_a$ & Similar to $N_w$, but refers to the number of different agents to ask for value-based and value-system based preferences over the previous $N_s$ pairs.\\
        $b_{ep}$ & Batch size of randomly selected pairs per agent from the preference replay buffer $R_p$ to obtain $C_V$ and $C_J$ for the E-step at each call to Algorithm~\ref{alg:algorithm1} in SVSL-P.\\
        $b_{mp}$ & Batch size of comparisons per agent from $R_p$ for M-steps at Algorithm~\ref{alg:algorithm1} in SVSL-P. \\
        $A_{ref}$ & Target representativeness and minimum coherence to achieve with Algorithm~\ref{alg:algorithm2} before starting the main loop in SVSL-P.\\
        $S_p$ & Maximum size of the preference replay buffer size $R_p$.\\
        $\gamma$ & Discount factor for preferences and RL policy learning. \\
        \bottomrule
    \end{tabular}
    \caption{Glossary of hyperparameters used in the experiments of SVSL-P and PbMORL.}
    \label{tab:hp-glossary-SVSLP-PbMORL}
\end{table}

The experiments where executed on a MacBook Pro with 16GB RAM, chip Apple M2. The code is not optimized for efficiency, as this was not in the scope of the paper. The current implementation of SVSL-P for the longest experiments ($L_{max}=15$, MVC environment) takes $7.1$ hours in average to run 5 seeds at once, with a $\pm 10$ minute difference across repetitions, due to the different neural network initializations that influence how soon a social value system is found with coherence and representativeness over the ``accuracy reference'' $A_{ref}$. 


\subsection*{Firefighters environment specification}\label{sec:env-specification}
In Table~\ref{tab:reward-firefighters} we specify the ground truth reward in the Firefighters MDP for each of the two values, given the current state, and action. The transition function is deterministic and is depicted in Table~\ref{tab:transition-firefighters}. The characteristics of the state space (determined by a series of categorical features) are presented in Table~\ref{tab:state-features-firefighters}. An important implementation detail is that we set a maximum environment horizon of 50 steps, after which any trajectory ends, but the Pareto efficient trajectories are typically of length 10, approximately, ending in states where the agent is able to put out the fire and rescue all the survivors.

\begin{table*}[h!]
    \centering
    \caption{Specification of actions and state features for the Firefighters MDP}
    \label{tab:state-features-firefighters}
    \begin{tabular}{p{0.3\linewidth}|p{0.6\linewidth}}
        \hline
        \textbf{Action} & \textbf{Description} \\
        \hline
        Evacuate Occupants &  Prioritize evacuating people from the building. \\
        Contain Fire &  Focus on containing the fire to prevent it from spreading. \\
        Aggressive Fire Suppression &  Engage in direct firefighting to reduce fire intensity quickly. \\
        Prepare Equipment &  Prepare equipment for safer execution of subsequent tasks. \\
        Update Knowledge & Update the knowledge about the fire and building status to plan subsequent actions. \\
        \hline
    \end{tabular}
    \begin{tabular}{p{0.23\linewidth}|p{0.25\linewidth}|p{0.416\linewidth}}
    \hline
        \textbf{State Feature} & \textbf{Possible Values} &\textbf{Description} \\
        \hline
        Fire Intensity & None, Low, Moderate, High, Severe & Severity of the fire at the current state. \\
        Occupancy & 0, 1, 2, 3, 4 & Area population density. \\
        Equipment Readiness & Not Ready, Ready & Availability of firefighting equipment. \\
        Knowledge & Poor, Good & Environmental condition affecting firefighting. \\
        Firefighter Condition & Perfect Health (3), Slightly Injured (2), Moderately Injured (1), Incapacitated (0) & Health status of the firefighter. \\
        \hline
    \end{tabular}
\end{table*}

\begin{table*}[h]
    \centering
    \caption{Firefighter Domain State-Transition Function. Legend: KN = Knowledge, FI = Fire Intensity, EQ = Equipment Readiness, OC = Occupancy, FFC = Firefighter Condition}
    \label{tab:transition-firefighters}
    \begin{tabular}{|l|l|l|}
        \hline
        \textbf{Action} & \textbf{State Conditions} & \textbf{State Transitions} \\
        \hline
        \multirow{3}{*}{Evacuate Occupants} 
            & Always & OC $\rightarrow \max(0, \text{OC} - 1)$ \\
            & FI $\geq 3$, EQ = 0, KN = 0 & FFC $\rightarrow \max(0, \text{FFC} - 1)$ \\
            & FI = 5 & EQ $\rightarrow 0$ \\
        \hline
        Contain Fire & Always & FI $\rightarrow \max(0, \text{FI} - 1)$ \\
        \hline
        \multirow{3}{*}{Aggressive Fire Suppression} 
            & Always & FI $\rightarrow \max(0, \text{FI} - 2)$ \\
            & FI $\geq 3$, (EQ = 0 $\vee$  KN = 0) & FFC $\rightarrow \max(0, \text{FFC} - 1)$ \\
            & FI = 5 & EQ $\rightarrow 0$ \\
        \hline
        Prepare Equipment & Always & EQ $\rightarrow 1$ \\
        \hline
        Update Knowledge & Always & KN $\rightarrow 1$ \\
        \hline
    \end{tabular}
\end{table*}

\begin{table*}[h]
    \centering
    \caption{Firefighter Domain Reward Specification. Legend: KN = Knowledge, FI = Fire Intensity, EQ = Equipment Readiness, OC = Occupancy, FFC = Firefighter Condition. To note is that if the firefighter is incapacitated in the next state, the rewards are $-1.0$ overriding any other situation. }\label{tab:reward-firefighters}
    \begin{tabular}{|l|l|c|c|}
        \hline
        \textbf{Action} & \textbf{State Conditions} & \textbf{Professionalism} & \textbf{Proximity} \\
        \hline
        \multirow{2}{*}{Evacuate Occupants} 
            & OC = 0 & -1.0 & -1.0 \\
            & OC $\neq$ 0 & $1 - 0.2 \times \text{FI} - 0.1 \times \text{KN}$ & 1.0 \\
        \hline
        \multirow{2}{*}{Contain Fire} 
            & FI = 0 & -1.0 & -1.0 \\
            & FI $\neq$ 0 & 0.8 & 0.2 \\
        \hline
        \multirow{3}{*}{Aggressive Fire Suppression} 
            & FI = 0 & -1.0 & -1.0 \\
            & FI $\neq$ 0, EQ = 0 & 0.3 & 0.7 \\
            & FI $\neq$ 0, EQ $\neq$ 0 & 0.6 & 0.7 \\
        \hline
        \multirow{2}{*}{Prepare Equipment} 
            & EQ = 0 & 0.5 & -0.1 \\
            & EQ $\neq$ 0 & -1.0 & -1.0 \\
        \hline
        \multirow{2}{*}{Update Knowledge} 
            & KN = 0 & 1.0 & -0.5\\
            & KN $\neq$ 0 & -1.0 & -1.0 \\
        \hline
        \multicolumn{2}{|c|}{Next state has FFC = 0 (Incapacitated)} & -1.0 & -1.0 \\
        \hline
    \end{tabular}
\end{table*}

\subsection*{Algorithms from previous works}\label{sec:algorithmssupp}
In this section we explain our approach used in the (to be published) previous work of ours~\cite{andresEcai2025} to solve Problem~5 of the main paper without the additional feedback and in non-sequential decision making. The algorithm here learns a grounding function and a value system of the society by observing pairwise comparisons between decision \textit{entities} (in our MDP case, these entities are trajectories) in terms of values and value systems of a given society of agents, but does not learn policies aligned with those values in the process. We employ these algorithms as a baseline in the evaluation and also as parts of the proposed SVSL-P algorithm. 

The main algorithm is based on EM (Expectation-Maximization) clustering (see Algorithm~\ref{alg:algorithm1}), which mimics~\cite{pmlr-v235-chakraborty24b}. There, the approach was used to learn a clustering of agents in terms of their preferences about pairs of options. To do so, it performs several times a cycle of two steps. In the first step, the algorithm assigns each agent to the cluster (a preference model) that better represents its preferences (E-Step, lines 3-5). In the second step (M-Step, Lines 6-16), the preference model of each cluster is trained to better fit the preferences of the assigned agents. 

The M-step from~\cite{pmlr-v235-chakraborty24b} consists on fitting a reward model $R^\theta(s,a)$ minimizing a loss similar to cross-entropy in the training data: 

$$\mathcal{L}\left(\tau,\tau',y\middle|R^\theta\right) = -y\log(p(\tau,\tau'|R^\theta)-(1-y)\log(p(\tau,\tau'|R^\theta))$$ 

In our case, we have to fit two groups of reward models (alignment functions). The first group is one model per value, i.e. the reward vector function $\Rv_V^\theta=\left(R_{v_1}^\theta,\dots, R_{v_m}^\theta\right)$; the second is composed by up to $L_{max}$ value system functions, that depend on the weights $W^\omega_l$, $l=1,\dots, L_{max}$ and the grounding models $\Rv_V^\theta$. 
Each group of models depend on different preferences in our dataset $DS$, which suggests two groups of loss functions, one based on the value system preferences $\mathcal{L}_{\VS}(DS|\beta)$, at Eq.~\ref{eq:loss_vs}, and another consisting of one loss per value of the grounding preferences $\mathcal{L}_{V}(DS)$, at Eq.~\ref{eq:loss_gr}.

\begin{align}\label{eq:loss_vs}
    &\mathcal{L}_{\VS}(DS|\beta) = \mathcal{L}_{\repr}(DS|\beta) - \mathcal{L}_{\conc}(DS|\beta), \\
    \label{eq:loss_vs1}&\mathcal{L}_\repr(DS|\beta) =\frac{1}{\abs{J}}\sum_{j\in J}\sum_{(\tau,\tau',y_V^j,...) \in DS_j}\frac{\mathcal{L}\left(\tau,\tau',y_V^j|R_{\beta(j)}^{\omega,\theta}\right)}{\abs{DS_j}}\\
    \label{eq:concisenessloss}&\mathcal{L}_{\conc}(DS|\beta) =\sum_{\substack{l \neq  l'}} \frac{1}{\abs{J}}\sum_{j\in J}\sum_{(\tau,\tau',...) \in DS_j}\frac{D(\tau,\tau'|R_l^{\omega,\theta},R_{l'}^{\omega,\theta})}{{\abs{DS_j}}}
\end{align}
and 
\begin{equation}\label{eq:loss_gr}
    \mathcal{L}_{V}(DS) = \left(\frac{1}{\abs{J}}\sum_{j\in J}\sum_{(\tau,\tau',...,y_{v_i}^j,\dots) \in DS_j} \frac{\mathcal{L}(\tau,\tau',y_{v_i}^j)|R_{v_i}^\theta)}{\abs{DS_j}}\right)_{i=1}^m
\end{equation}
where 
\begin{align}
    &\mathcal{L}\left(\tau,\tau',y|R\right) = -y\log(p(\tau,\tau'|R))-(1-y)\log(p(\tau,\tau'|R)),\label{eq:cross-entropy}\\ 
    &\text{where: }
     p(\tau,\tau'|R) = \frac{\exp{ \sum_{(s,a)\in\tau} R}}{\exp{\sum_{(s,a)\in\tau} R(s,a)}+\exp{ \sum_{(s,a)\in\tau'} R}}\label{eq:bradley-terry2}
\end{align} 

Our ``value system loss'' in Eq.~\ref{eq:loss_vs} has two terms. The first term, in Eq.~\ref{eq:loss_vs1} increments representativeness by minimizing discordance (Eq.~3 from main paper). The second term (Eq.~\ref{eq:concisenessloss}) increases conciseness by separating the preference models of the most similar clusters. 

As conciseness is not differentiable, we employ a quantitative version of inter-cluster discordance, using the term $D(\tau,\tau'|R_1,R_2)$: the Jensen Shannon Divergence between the Bernoulli distributions of parameters $p(\tau,\tau'|R_1)$ and $p(\tau,\tau'|R_2)$. Incrementing this metric tends to increase the discordance between $\preccurlyeq_{R_1}$ and $\preccurlyeq_{R_2}$, which increases conciseness. {Jensen-Shannon divergence has also been used in the related problem of finding the centroid of probability distributions~\cite{jensenfordistancebetweenprobscentroid}}.



The grounding loss for each value $v_i$, with $i=1,\dots, m$, (Eq.~\ref{eq:loss_gr}) is a cross-entropy loss computed over the dataset, aggregating the examples of each agent separately. Minimizing these losses increases grounding coherence by reducing discordances.  

The grounding and value system loss functions need to be minimized in a hierarchy, i.e., prioritizing the grounding loss to improve coherence, and in second place, consider the value system loss. We approach this as a constrained optimization problem. The constraints to satisfy here are maximizing the coherence with each value, i.e., finding grounding network parameters $\theta$ such that $ \chr_{D_{J}}(\preccurlyeq_{R^{\theta}_{v_i}}) = \chr^*_{v_i}$, with $\chr^*_{v_i} = \max_{\theta \in \Theta}\chr_{D_{J}}(\preccurlyeq_{R^{\theta}_{v_i}})$, for every $i\in \{1,\dots, m\}$. Since $\chr^*_{v_i}$ is unknown a priori, it is dynamically estimated as the highest coherence observed during the learning process (smoothed by an exponentially weighted maximum: Algorithm~\ref{alg:algorithm1}, Line 16). The constraint to satisfy in terms of our loss function should be $\mathcal{L}_V(DS) \leq \mathcal{L}^*_{V}$, where $\mathcal{L}^*_V$ is a loss that guarantees maximum coherence with all values. As we do not know $\mathcal{L}_V^*$, we assume the stricter constraint $\mathcal{L}_V(DS) = 0$. With $m$ positive Lagrange multipliers $\lambda = (\lambda^1, \dots, \lambda^m)$ our objective is transformed to:

\begin{align}\label{eq:Lagrange}
    \min_{\theta,\omega} \max_\lambda\mathcal{L}_{\VS}(DS|\beta) &- \sum_{i=1}^m\lambda_i\cdot  \left(\mathcal{L}_V(DS)\right)
\end{align}

We seek a Nash equilibrium of jointly minimizing the Lagrangian in Eq.~\ref{eq:Lagrange} over $\theta,\omega$ (subject to the assignment $\beta$) and maximizing over $\lambda \in [\R^+]^m$~\cite{pmlrv98cotter19aLagrangianNash}. This is done through successive iterations of improving the Lagrangian (via gradient descent: Algorithm~\ref{alg:algorithm1}, Line 10) and then increasing the Lagrange multipliers $\lambda$ through gradient ascent with a learning rate $\alpha_\lambda$ (Algorithm~\ref{alg:algorithm1}, Line 14). To avoid overfitting the artificial constraint $\mathcal{L}_V(DS) = 0$, the Lagrange multipliers for each value $v_i$ increase only when the coherence is below $\chr^*_{v_i}$, and only for the value that is the farthest from its maximum found coherence.  Furthermore, multipliers are decayed using a factor $\gamma_\lambda$ if coherence remains at $\chr^*_{v_i}$. Also, as we use small batches for training (differently from~\cite{andresEcai2025}), to not overestimate $\chr^*_i$ from ``lucky'' batches, we approximate it with an exponentially weighted \textit{maximum}, i.e. if $\chr_{B}(\preccurlyeq_{\Rv_V}) \geq \chr^*_i$, we update the maximum coherence with $\chr^*_i = r_\lambda\chr^*_i + (1-r_\lambda) \chr_{B}(\preccurlyeq_{\Rv_V})$ where $B$ is a certain batch of compared trajectory pairs from the whole society and $r_\lambda$ is set to a number close to $1$. A good value for $r_\lambda$ was found to be $0.99$ across the tested environments.

EM algorithms are known to converge to local optima or stationary points~\cite{emlocalconvergence}, depending on initialization. To address this, Algorithm~\ref{alg:algorithm2} (which constitutes the SVSL algorithm from the paper) introduces an exploitation-exploration outer loop inspired by evolutionary algorithms (EA), extending the EM procedure in Algorithm~\ref{alg:algorithm1}. A memory $M$ (that acts as the EA \textit{population}) of social value systems is kept. At each iteration, a solution is selected from $M$ based on its quality (Line 5), mutated with probability $\text{\textit{mrt}} > 0$ (Line 7), and then refined with Algorithm~\ref{alg:algorithm1} (Line 9) during $E_r$ epochs where the first cycle directly performs the M-step over the mutated solution (Algorithm~\ref{alg:algorithm1}, Line 3 ). Finally, it returns a new social value system.

The new solution is inserted in the memory (Algorithm~\ref{alg:algorithm2}, Line 10), replacing an existing one if it Pareto-dominates it. Pareto dominance is based on grounding coherence, number of clusters, conciseness, and representativeness. The memory has a capacity $N$, requiring an elimination protocol under overflow (Line 11). We seek a balance between keeping quality solutions --according to coherence, Ray-Turi Index and Pareto dominance-- for exploitation, and maintaining varied clusterings for exploration. The eliminated solution is chosen as the worst in the following lexicographic order: (1) higher number of clusters, (2) number of identical agent-cluster mappings, (3) number of dominating solutions, (4) grounding coherence, and (5) lower Ray-Turi index. Solutions with the best coherence and Ray-Turi index are always preserved. 

The selection step (Algorithm~\ref{alg:algorithm2}, Line 5) involves, first, ordering the options by the outer optimization objective (Ray-Turi Index), and then, by the inner objective (grounding coherence). This order inversion is intentional, as coherence can in all cases be improved via the Lagrange multiplier method, while Ray-Turi Index, and in particular, conciseness is best improved through exploration. Then, a solution is chosen with probability proportional to its rank (following Eq.~2 from~\cite{linearRankSelectionEq2}). 

The mutation step (Line 7) involves two tasks. First, it either removes a certain number of clusters (redistributing its agents randomly) or adds a certain number of them, populated by moving agents to these with a probability $p_m$. Second, it perturbs the parameters of both the grounding network and value system weights using Gaussian noise, following classical evolutionary strategies~\cite{fogelevolutionary}. The magnitude of perturbation is scaled by the coherence error for $ \theta$ and the representativeness error for $\omega$\footnote{``Error'' in this context is understood as the difference of the learned coherence and representativeness to their maximum, i.e. the ideal value of $1.0$.}.

\begin{algorithm}[t]
\raggedright
\caption{Value System Learning of a Society (EM algorithm)}\label{alg:algorithm1}
\textbf{Initialization:} Dataset $DS$ composed of trajectory pairs $D$ compared by different agents (in terms of value alignment with each value in a set $V$ and in terms of their value system), i.e. $DS=\cup_{j\in J}DS_j$, $D=\cup_{j\in J}D_j$. Learning rates $\alpha_\theta$, $\alpha_\omega$, $\alpha_\lambda$. Lagrange multiplier decay $\gamma_\lambda > 0$. Maximum number of clusters $L_{max}$. Number of M-Step repetitions ($m_r$), number of EM epochs $E_r$. Set maximum achievable coherence $\chr^*_{v_i} = 0$ for all $i$.

\textbf{Input:} 
Assignment $\beta$ (optional), parameters of the value system weights $\omega$, parameters of the grounding network $\theta$, Lagrange multiplier state $\lambda = \left(\lambda_1,\dots,\lambda_m\right)$ (optional, otherwise use initialization). Exponential weighted maximum coefficient to update coherences $r_\lambda$. Optionally, additional preference replay buffer $R_p$ and batch size per agent for the E-step $b_{ep}$ and per M-step repetition $b_{mp}$

\textbf{Output:} An assignment of agents into clusters $\beta$, updated parameters $\theta$, $\omega$ and new Lagrange multipliers $\lambda$.

 \begin{algorithmic}[1]
    \State Set $\Rv^\theta$ and $W_l^{\omega}$ (for $l\in \{1,\dots, L_{max}$\}) with params. $\theta$ and $\omega$, resp.
    \For{epoch $r=0,\dots,E_r-1$}
    \State \textproc{\textbf{E-Step}} (if $R_p$ supplied, for each agent $j$, sample a set of trajectory pairs $O_j$ from it of size $b_{ep}$:
        \State $\beta(j) \gets \argmin_{l} d_{D_j\cup O_j}\left(\preccurlyeq_{\Rv^\theta}^{W_l^\omega}, \preccurlyeq_V^j\right)$ \Comment{Do for all $j\in J$}
        \State \textbf{Simplify clusters (New from~\cite{andresEcai2025})}: for each pair $l,l'<L_{max}$, if $\max(|W_l-W_{l'}|)<0.01$, merge clusters to the one with most agents. Put random weights in the other.
    \State \textproc{\textbf{M-Step}} 
    \For{\textit{m-step} $t=0,\dots,m_r$} \State $BS \gets$ batch of $b_{mp}$  traj. pairs per agent from $R_p$
    \State $\mathcal{L}_{global} =  \mathcal{L}_{\VS}(BS|\beta) + \lambda\cdot  \left(\mathcal{L}_V(BS)\right)^T$
    \State $\theta \gets \theta - \alpha_\theta\nabla_\theta\mathcal{L}_{global}  $; $\omega \gets \omega - \alpha_\omega\nabla_\omega\mathcal{L}_{global}  $ 

    \EndFor
    \State Compute $C_i = \chr^*_{v_i} - \chr_{B}(\preccurlyeq_{R_{v_i}^\theta})$ for each $i=1,\dots,m$. $i* = \argmax C_i$
    \If{$C_i$ is maximum}
    \State $\lambda_{i^*} \gets (1-\gamma_\lambda)\lambda_{i^*} + \alpha_{\lambda} \left(\mathcal{L}_{V}(BS)\right)_i$
    \EndIf
    
    \State $\chr^*_{v_i} \gets (1-r_\lambda)\max\left[\chr_{B}\left(\preccurlyeq_{R_{v_i}^\theta}\right), \chr^*_{v_i}\right] + r_\lambda \chr^*_{v_i}$
    \EndFor
    \State \textbf{Return} $\beta$, $\omega$, $\theta$, $\lambda$
\end{algorithmic}
\end{algorithm}

\begin{algorithm}[h]
\caption{Value System Learning of a Society (\textbf{SVSL})}\label{alg:algorithm2}
\textbf{Input:}. All the initialization parameters from Algorithm~\ref{alg:algorithm1}. Number of training steps $I$. Memory of candidate solutions size $N$. Epochs per training step, $E_r$. Mutation probability $\textit{mrt} < 1$, agent reassignment probability $p_m$, network parameter mutation scale $s_m$. Initial Lagrange multipliers $\lambda_=  \left(\lambda_1,\dots,\lambda_m\right)$, $\lambda_i > 0$.

\textbf{Output:} An assignment of agents into clusters $\beta$, and trained reward vector $\Rv^\theta$ and value system weights $\pmb{W}^\omega = \left(W^\omega_{1},\dots,W^\omega_{L_{max}}\right)$ (from which only the $L$ weights corresponding to the clusters selected through $\beta$ are used).

 \begin{algorithmic}[1]
    \State Initialize Algorithm~\ref{alg:algorithm1}
    \State Generate value system network parameters $\omega$ and grounding parameters $\theta$;
    \State Repeat Line 2 $N$ times to fill memory $M$ (add multipliers $\lambda$).
    \For{training step $t = 0,\dots, I-1$}
    \State $\beta,\theta,\omega,\lambda \gets$\textproc{SelectSolution}($M$)
    \If{Rand$() < \text{\textit{mrt}}$}
    \State $\beta,\theta,\omega$  $\gets$ \textproc{MutateSolution}($M$, $p_m$, $s_m$)
    \EndIf
    
    \State $\beta',\theta',\omega',\lambda'\gets$\textproc{\textbf{Algorithm~\ref{alg:algorithm1}}}($\beta, \theta,\omega,\lambda$) with data in $DS$    
    \State \textproc{InsertInMemory($\beta',\theta',\omega',\lambda'$, $M$)}
    \State If $M$ is full: \textproc{EliminateWorstSolution(M)}
    \State Decrease \text{\textit{mrt}} linearly upto $0$ when $t=I-1$.
    \EndFor
     
 \State $\beta$,  $\pmb{W}^{\omega}$,$\pmb{R}^\theta$, $\lambda$ $\gets$ \textproc{GetBestSolution(M)}
 \State \textbf{return} $\beta$, $\pmb{W}^{\omega}$, $\Rv^\theta$, $\lambda$
 \end{algorithmic}
\end{algorithm}

\subsection*{Implementation details}

We introduce some new implementation details over the previous algorithms with respect our previous work~\cite{andresEcai2025}.

First, 
we regularize the basic cross-entropy loss in Eq~\ref{eq:cross-entropy} with label smoothing (set to $0.1$) to avoid overfitting the qualitative labels $y\in \{0,0.5,1\}$. This converts the labels to $y\in \{0.1,0.5,0.9\}$. We also use the Adam optimizer with a weight decay factor of $10^{-4}$. In Eq.~\ref{eq:concisenessloss}, we make the loss of the conciseness the sum over the Jensen Shannon divergences (the term $D(\tau,\tau'|R_l^{\omega,\theta},R_{l'}^{\omega,\theta})$, see~\cite{andresEcai2025}) between used clusters, instead of the minimum, to smooth the optimization process.  


Lastly, we introduce a mechanism to avoid redundant clusters during the learning process, at the end of the \textit{E-step} of Algorithm~S1, supplementary material, by simply ``merging'' clusters with ``close'' value system weights. Specifically, we consider two clusters equivalent when the maximum distance between the weights for any particular value is less than certain amount. We set this amount to $0.05$. Then, we send all the agents to the cluster with highest population among each pair of clusters that are considered as equivalent.

\subsection*{Additional results}

In Figure~\ref{fig:paretoeqlffPbMORL} (for PbMORL) and Figure~\ref{fig:paretoeqlffSVSLP} (for SVSL-P) we represent, for the ten seeds used, the full learned Pareto fronts and the expected value alignment of the clusters learned by each algorithm to represent the value system of the respective societies. We see that the Pareto fronts by PbMORL are slightly less close to the ground truth than those of SVSL-P, but that can be due to the different convergence rates. What is clear is that the clusters considered after learning with PbMORL are not diverse enough to cover the full front, while the ones by SVSL-P are consistently well distributed.

Seed variability in the number of clusters is notable in all cases. By contrast, the standard deviations in representativeness and coherence are very small. Our algorithm seems to find a way to maximize representativeness and coherence (as we stated in our problem formulation),  but does not consistently minimize cluster size (which is stated only indirectly with the conciseness goal). We suspect that cluster size variability might be due to our selected environments (specially FF), where different weight combinations in the front (value systems) actually represent very similar or equal qualitative preferences over the available trajectory spaces. Thus, selecting apparently different cluster combinations (and weights) does not always affect the qualitative preference errors (discordances) that our algorithm is designed to minimize (not the number of clusters directly). Furthermore, it seems clear that EQL tends to converge early in PbMORL, Figure~\ref{fig:paretoeqlffPbMORL}. Probably increasing the selection of ``disrupting'' preferences among the clusters would help to make the optimization aware of the cluster relevance. Another tentative patch to this seed variability would be employing \textit{ensembles} of reward functions~\cite{christiano2023deeprlpreferences}.

In both cases we observe that, occasionally, two or more clusters have their respective policies ``assigned'' to the same point on the front, or, as noted before, there is a cluster that is slightly less efficient than the policy learned for a value system that was \emph{not} selected to be a cluster. These issues merit attention in future work and raise two questions: (i) whether it is feasible to enforce that two clusters whose policies “close” or equivalent in the objective space should be conjoined, even when qualitative differences justify the existence of both clusters; and (ii) whether it is possible to focus the reinforcement learning process on the policies of the selected clusters only (or to do so with stronger efficiency guarantees) and what are the trade-offs with our current method (that explores the whole simplex weight space with gaussian probability).

\begin{figure*}[h]
    \centering\includegraphics[width=0.33\linewidth,trim=0.1cm 0.2cm 0cm 1cm, clip]{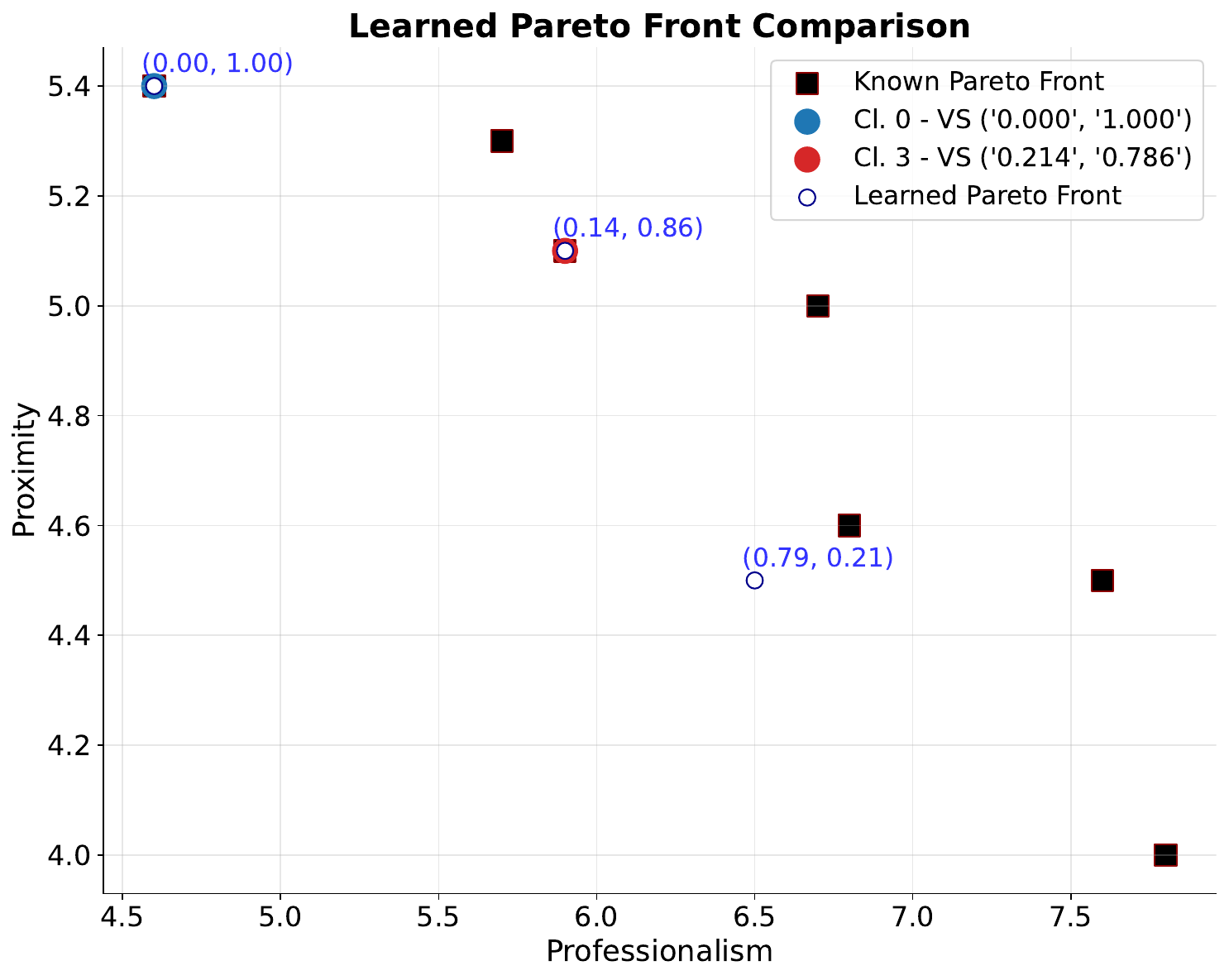}
    \includegraphics[width=0.33\linewidth,trim=0.1cm 0.2cm 0cm 1cm, clip]{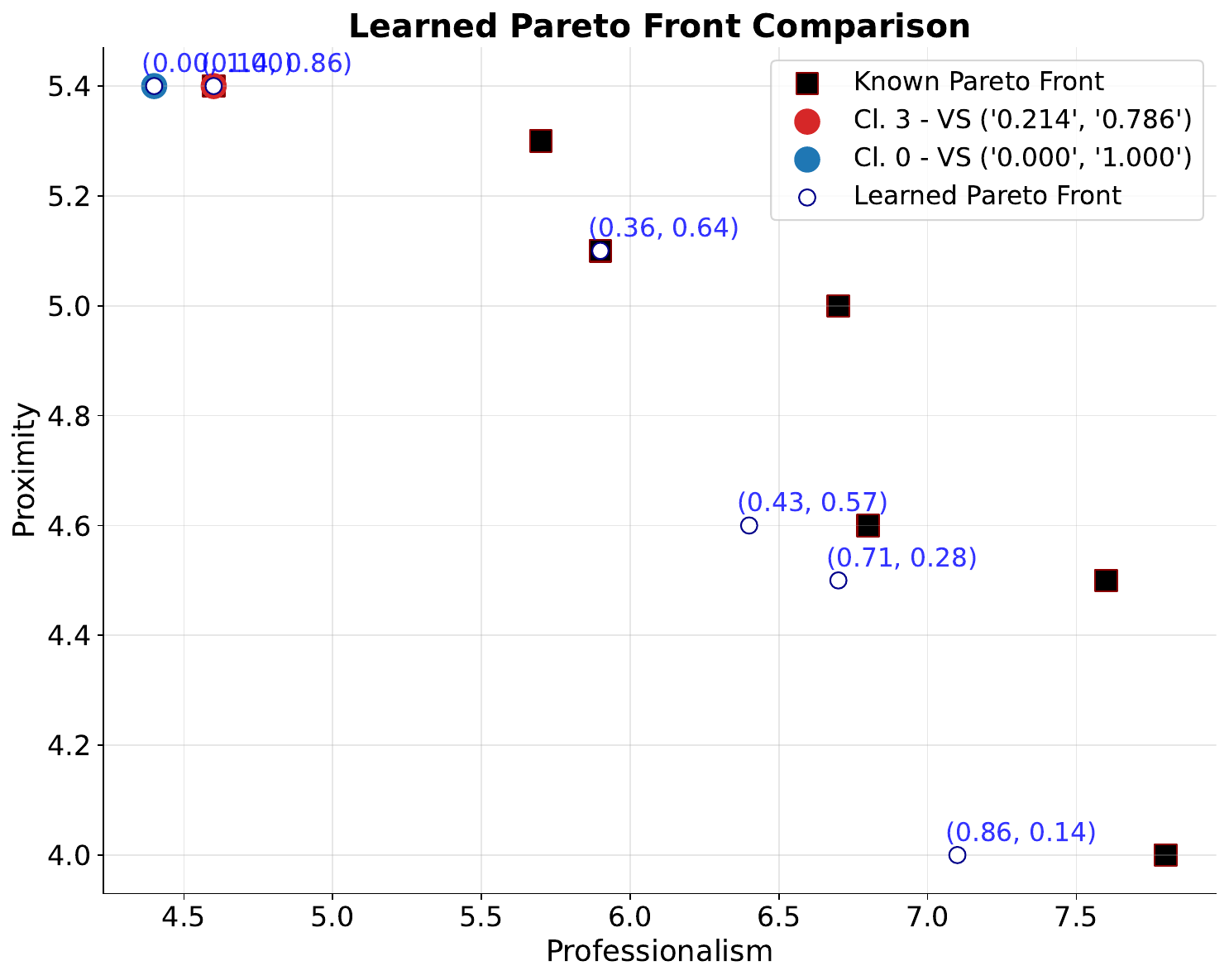}
    \includegraphics[width=0.33\linewidth,trim=0.1cm 0.2cm 0cm 1cm, clip]{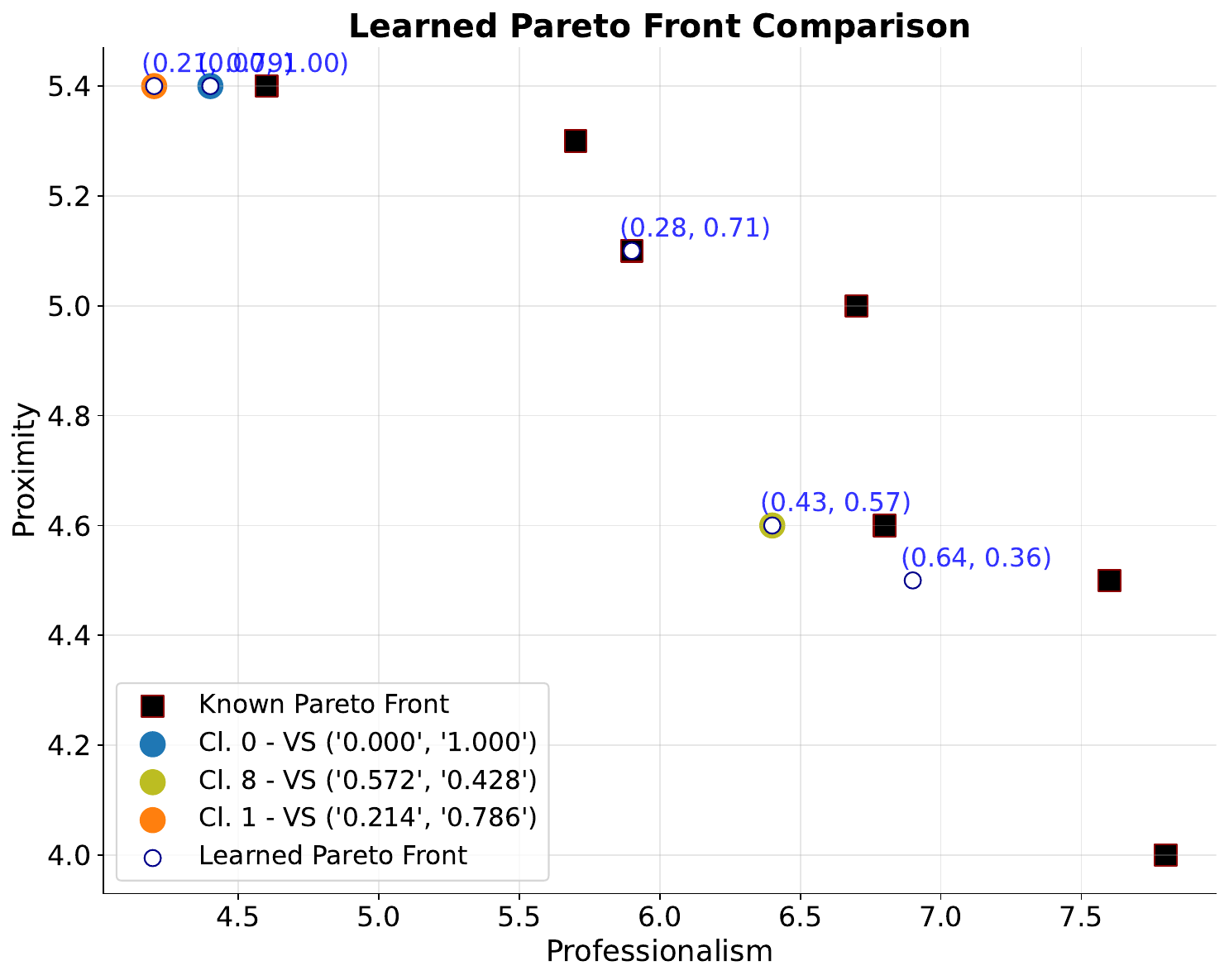}\includegraphics[width=0.33\linewidth,trim=0.1cm 0.2cm 0cm 1cm, clip]{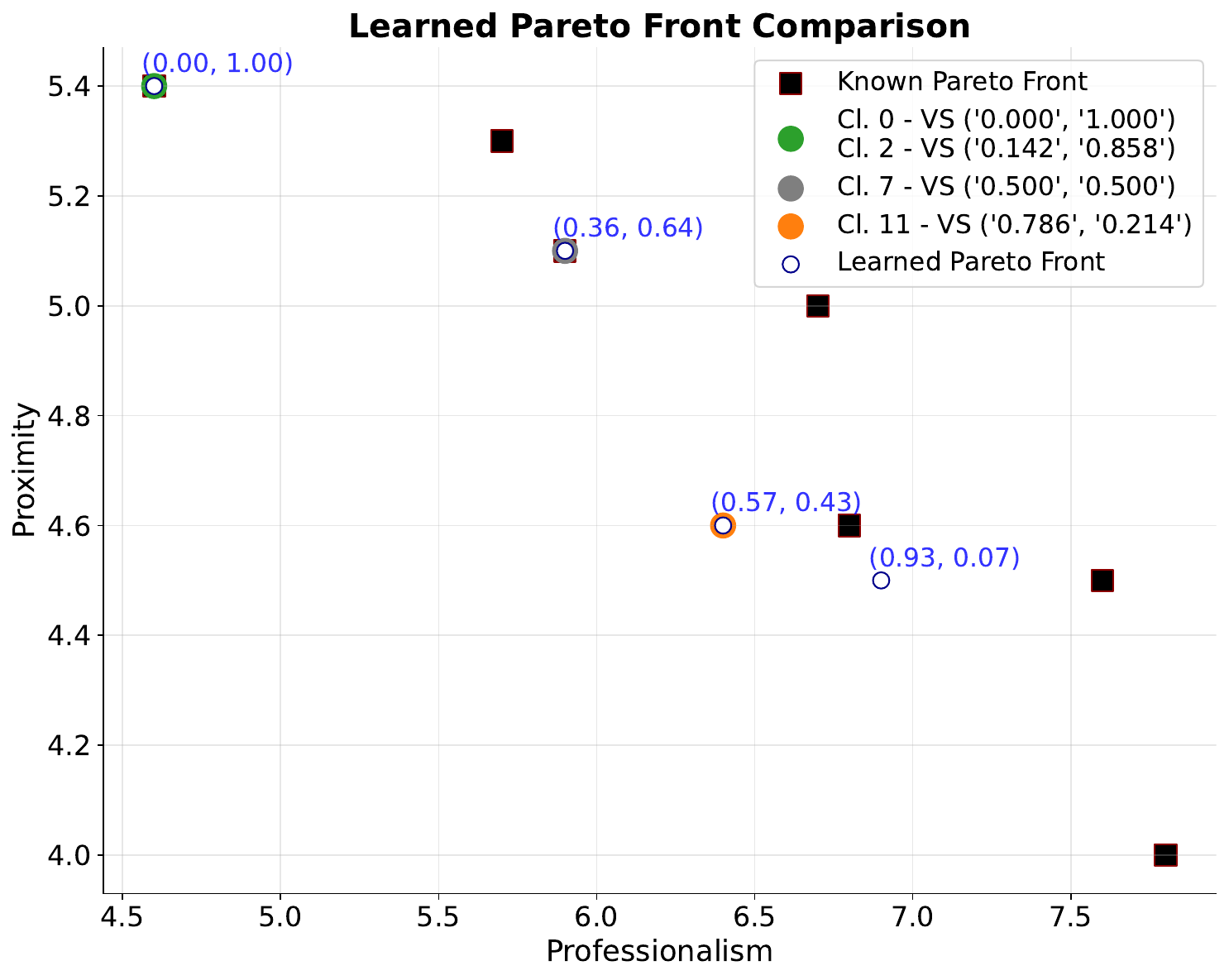}
    \includegraphics[width=0.33\linewidth,trim=0.1cm 0.2cm 0cm 1cm, clip]{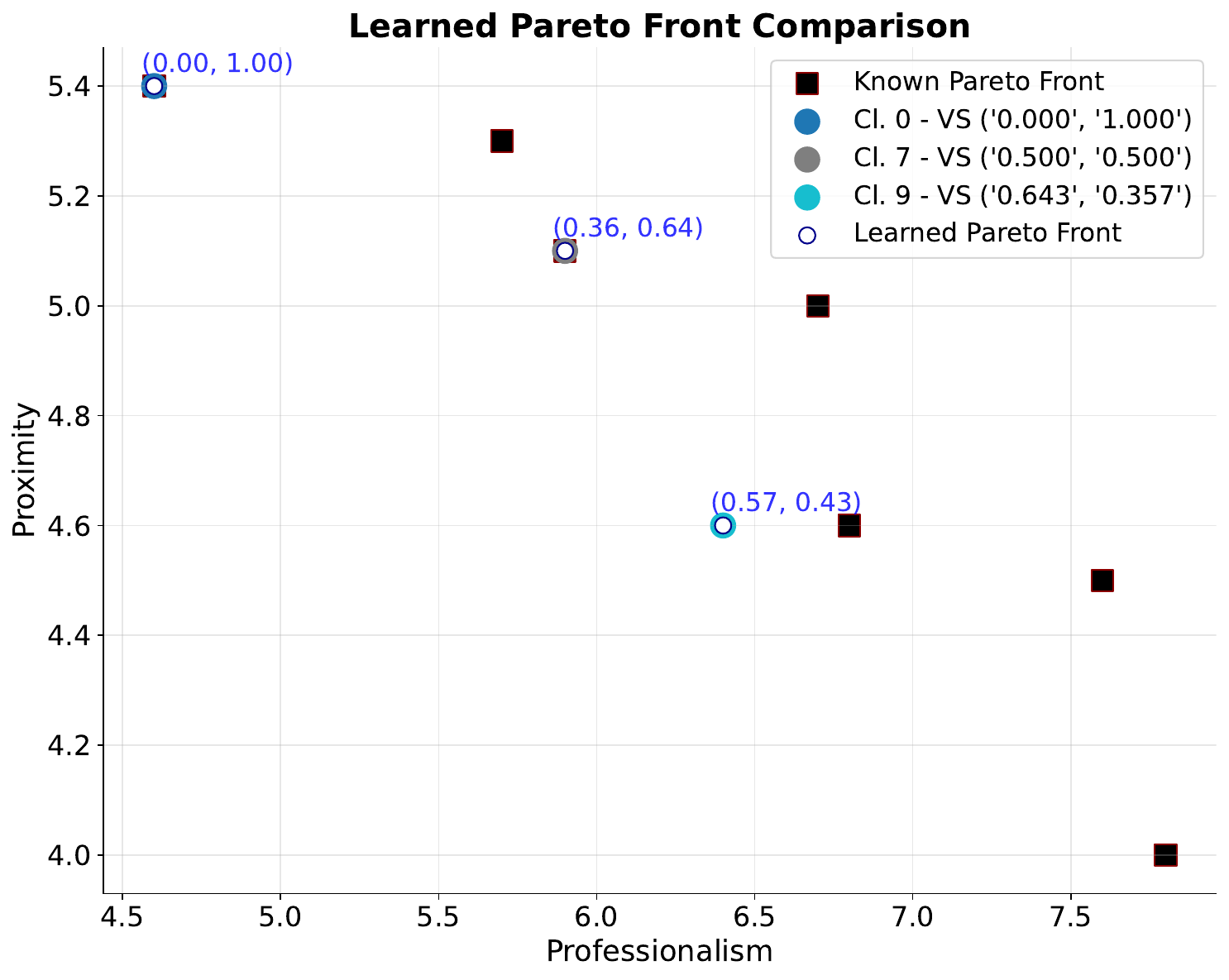}
    \includegraphics[width=0.33\linewidth,trim=0.1cm 0.2cm 0cm 1cm, clip]{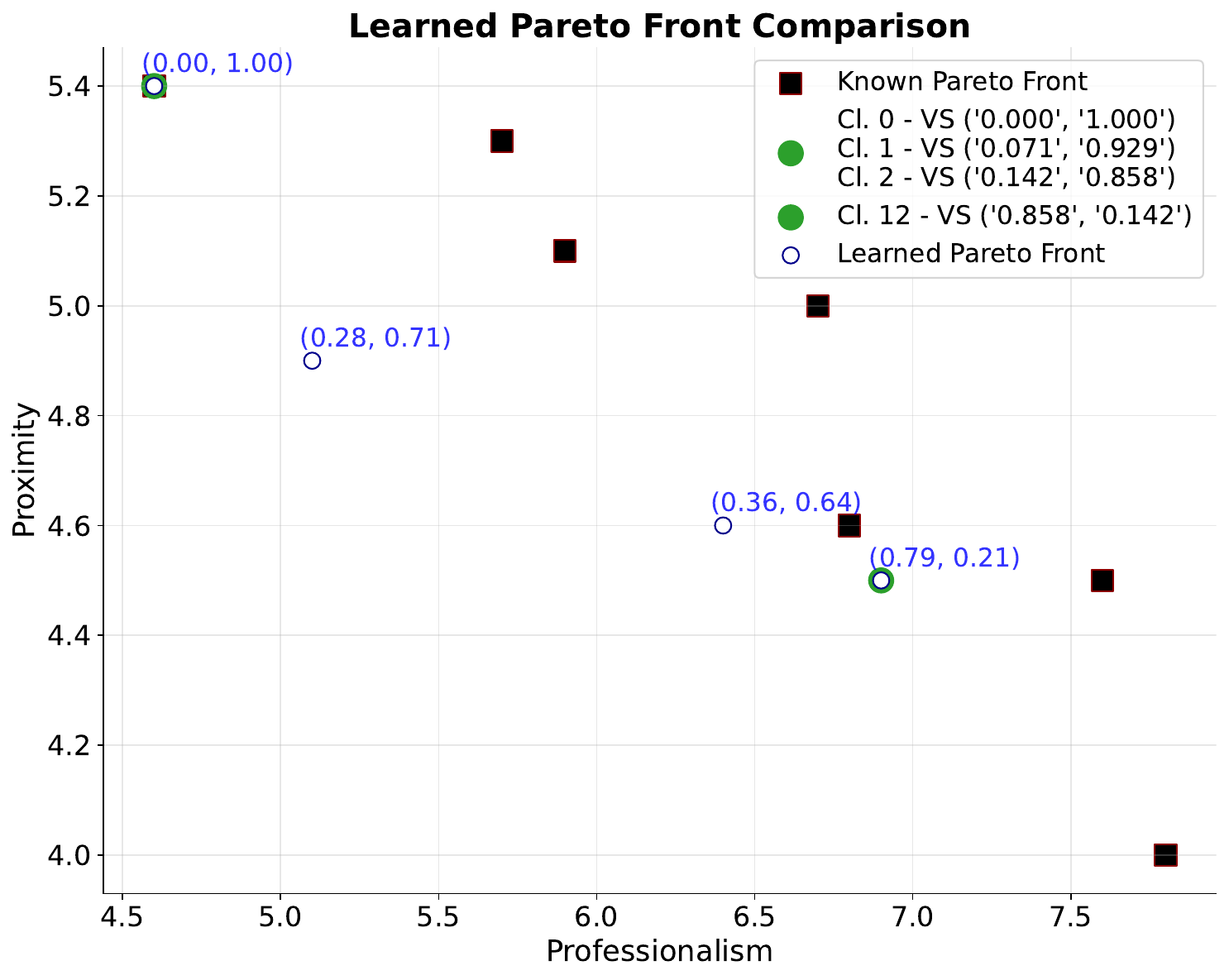}
    \includegraphics[width=0.33\linewidth,trim=0.1cm 0.2cm 0cm 1cm, clip]{media/ff/pareto/pbmorl/seed_31.pdf}
    \includegraphics[width=0.33\linewidth,trim=0.1cm 0.2cm 0cm 1cm, clip]{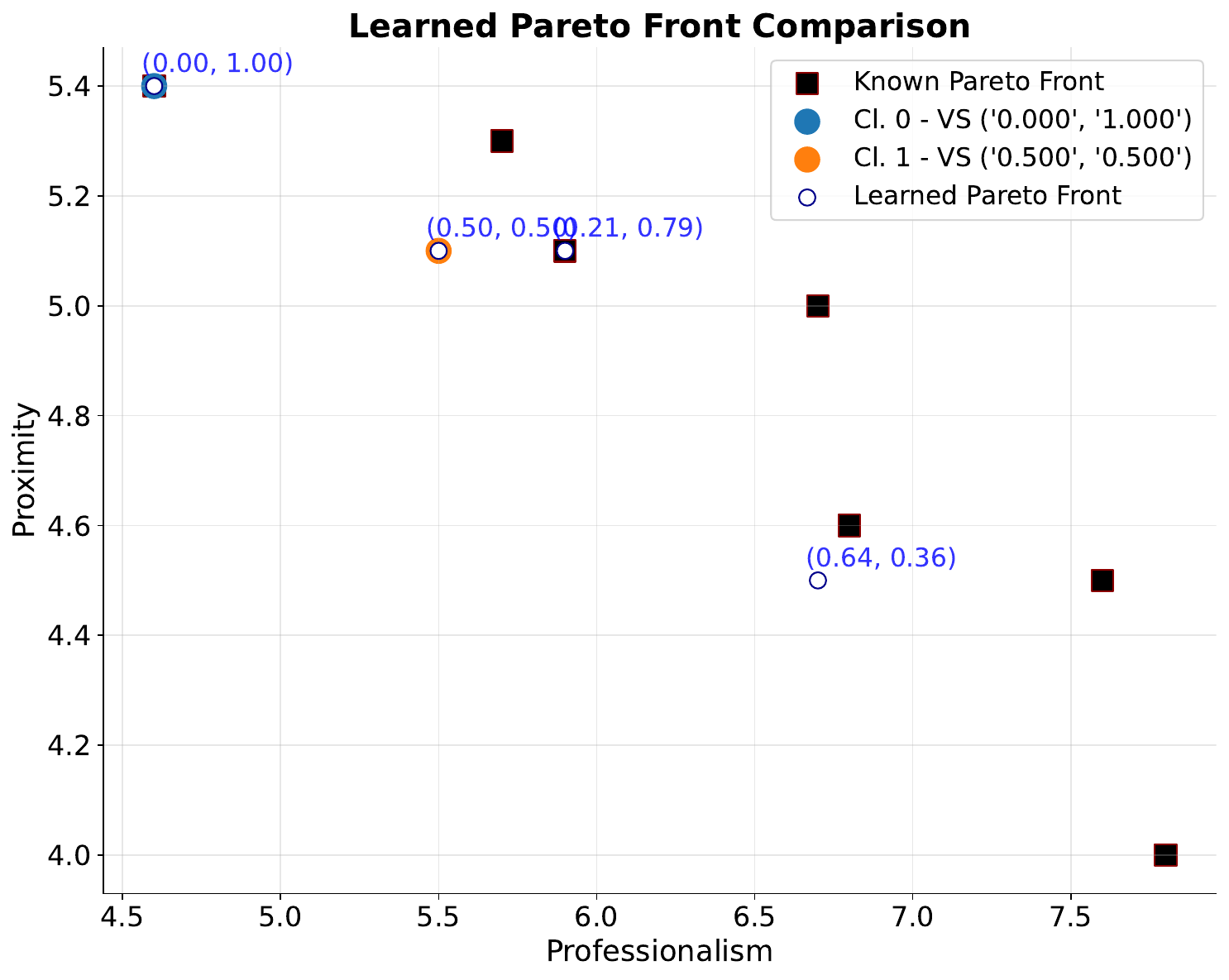}
    \includegraphics[width=0.33\linewidth,trim=0.1cm 0.2cm 0cm 1cm, clip]{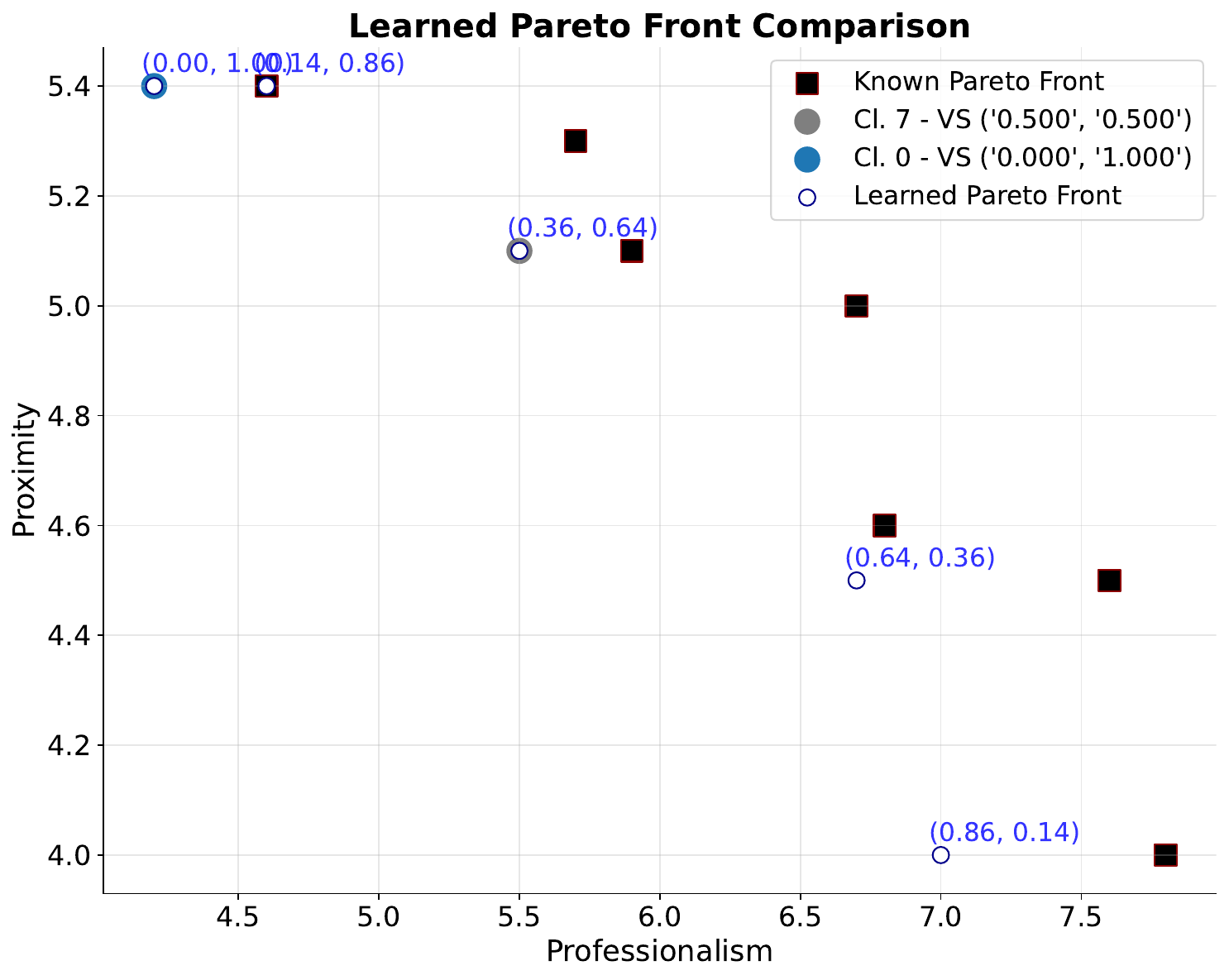}
    \includegraphics[width=0.33\linewidth,trim=0.1cm 0.2cm 0cm 1cm, clip]{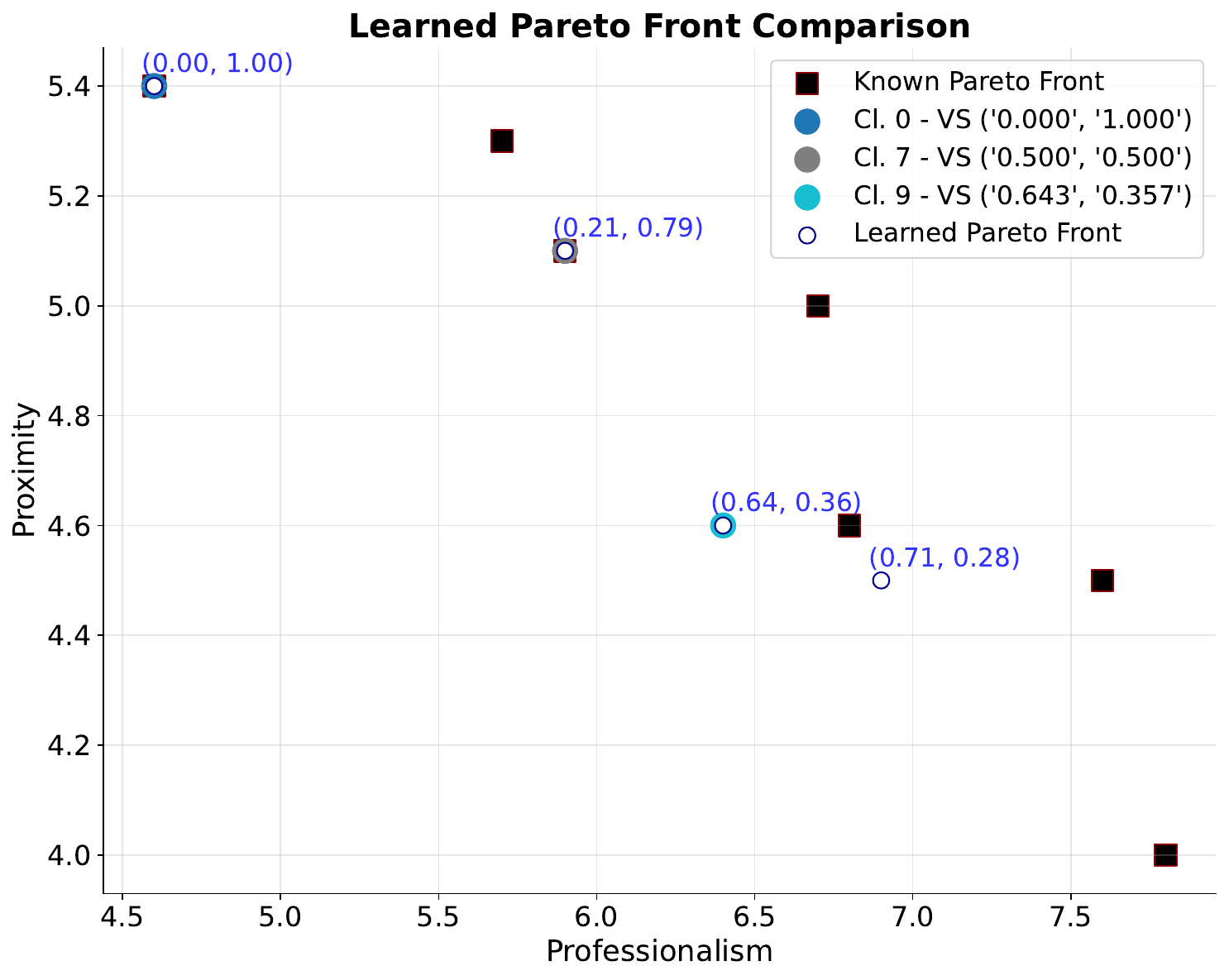}
    
    \caption{FF environment. Pareto front and clusters learned with PbMORL with the different 10 seeds. Black squares indicate the known Pareto front of the environment in terms of the alignment with the two values. White dots depict weights which policies are in the learned front with each method. Coloured white dots indicate the value system weights identifying each learned cluster (in the legend). Note that not all of the latter are necessarily efficient. }
    \label{fig:paretoeqlffPbMORL}
\end{figure*}

\begin{figure*}[h]
    \centering

    \includegraphics[width=0.33\linewidth,trim=0.1cm 0.2cm 0cm 1cm, clip]{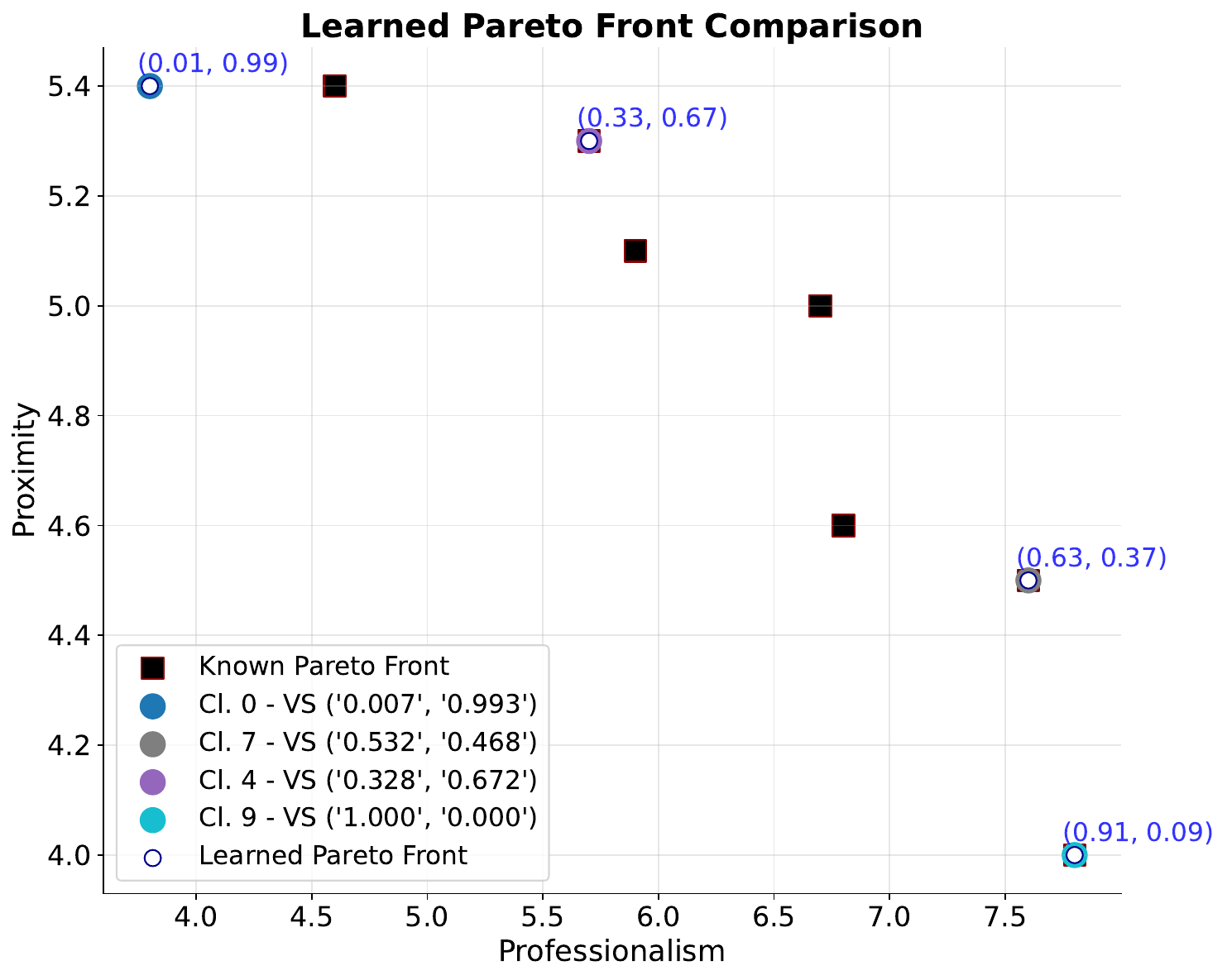}
    \includegraphics[width=0.33\linewidth,trim=0.1cm 0.2cm 0cm 1cm, clip]{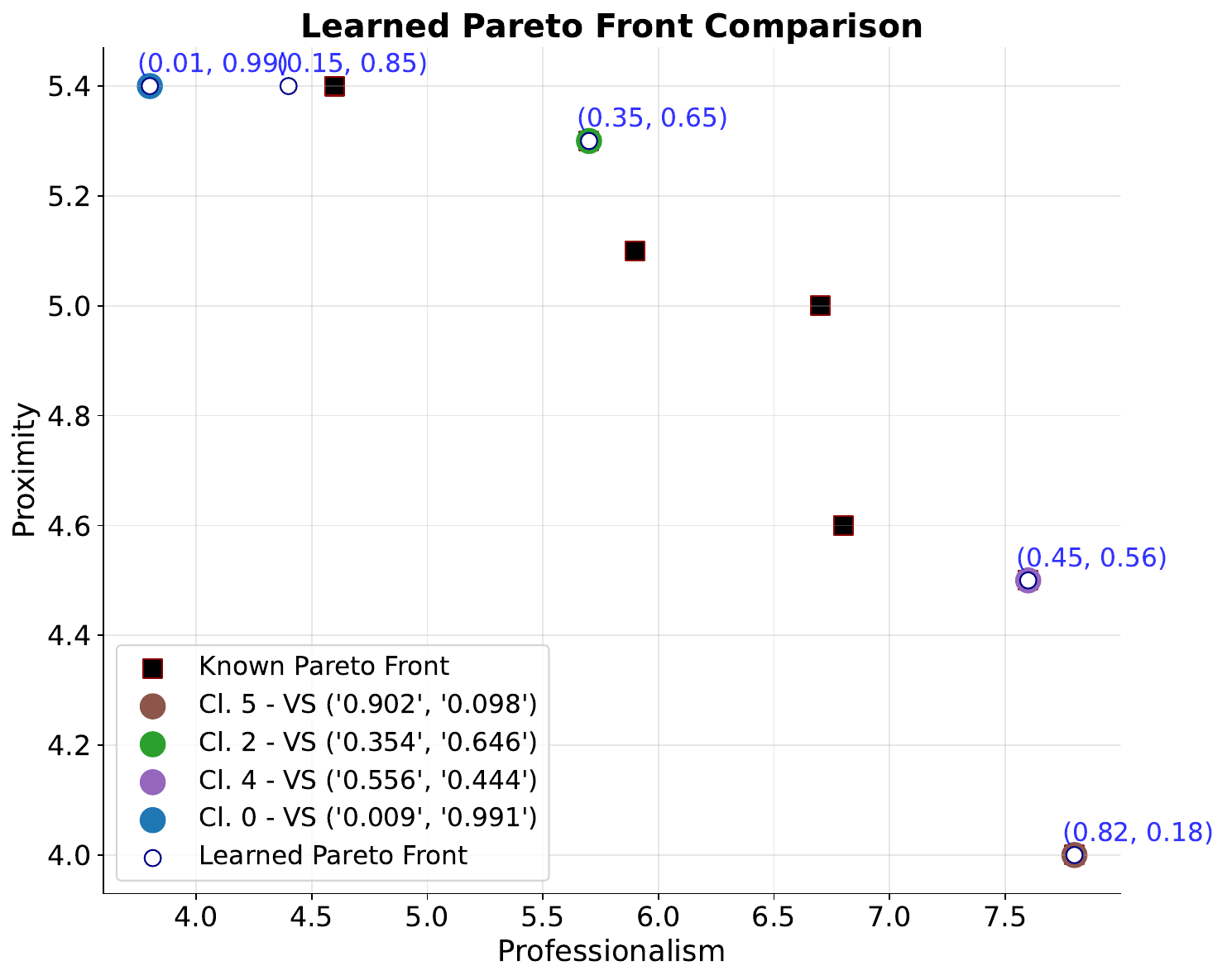}
    \includegraphics[width=0.33\linewidth,trim=0.1cm 0.2cm 0cm 1cm, clip]{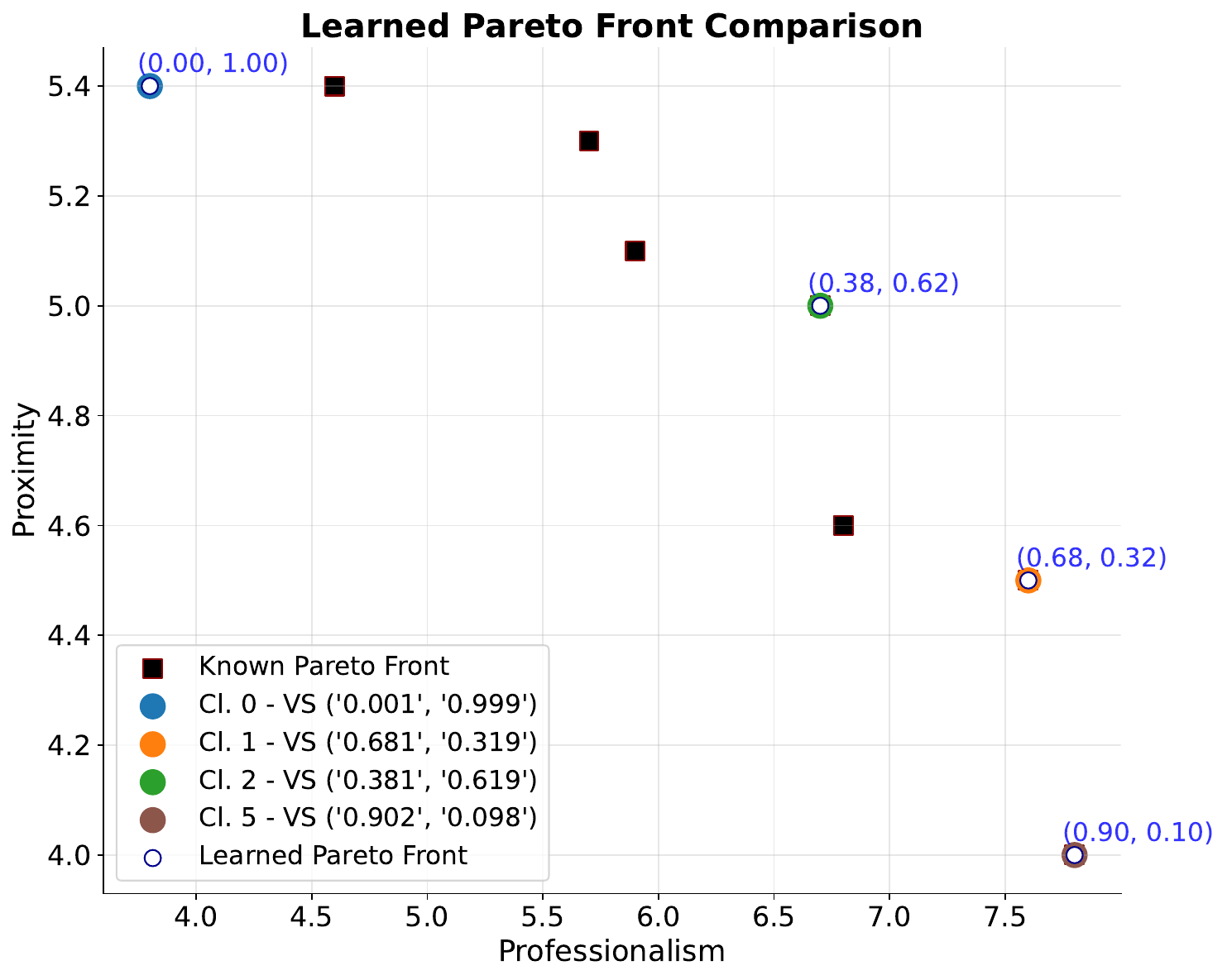}
    \includegraphics[width=0.33\linewidth,trim=0.1cm 0.2cm 0cm 1cm, clip]{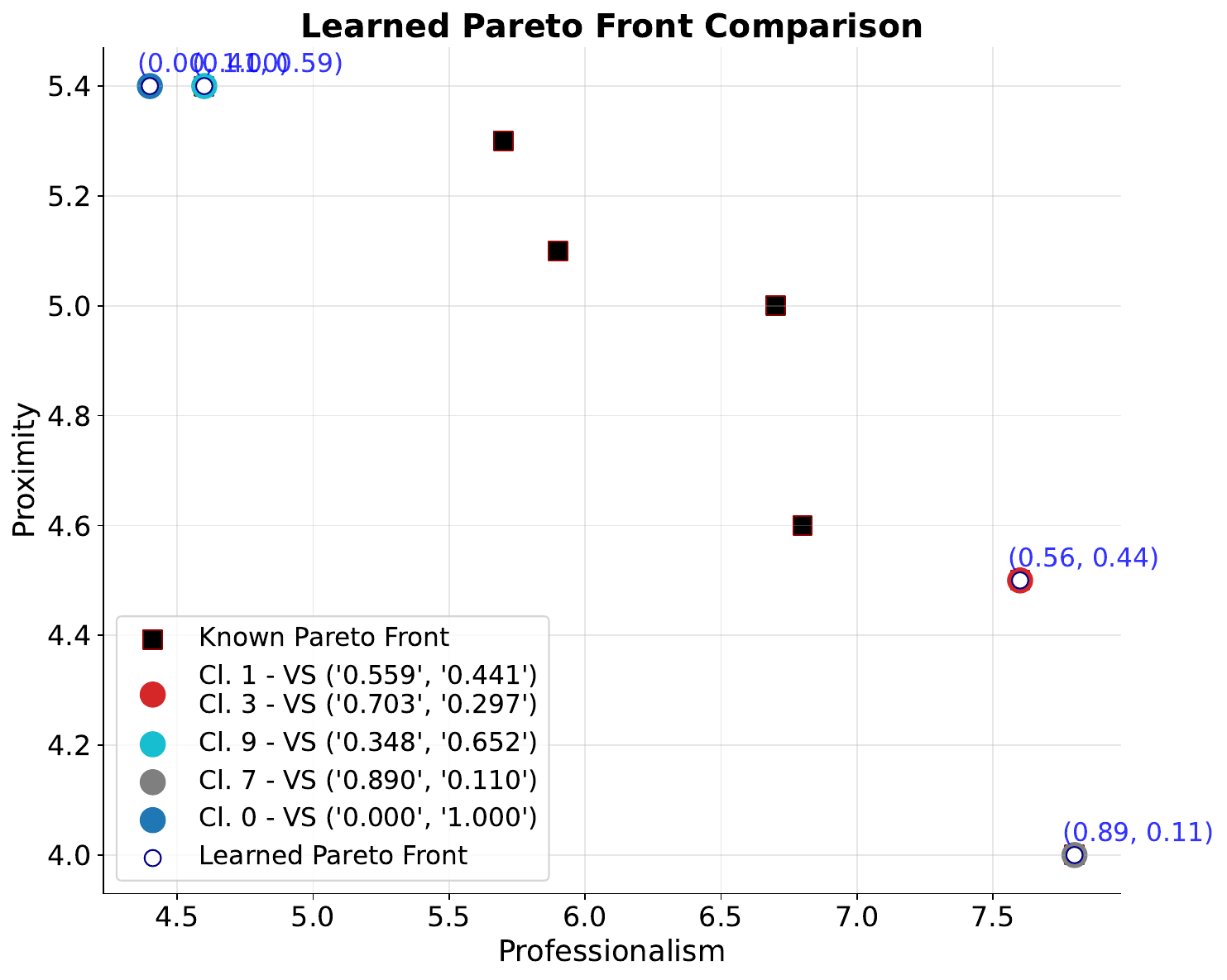}
    \includegraphics[width=0.33\linewidth,trim=0.1cm 0.2cm 0cm 1cm, clip]{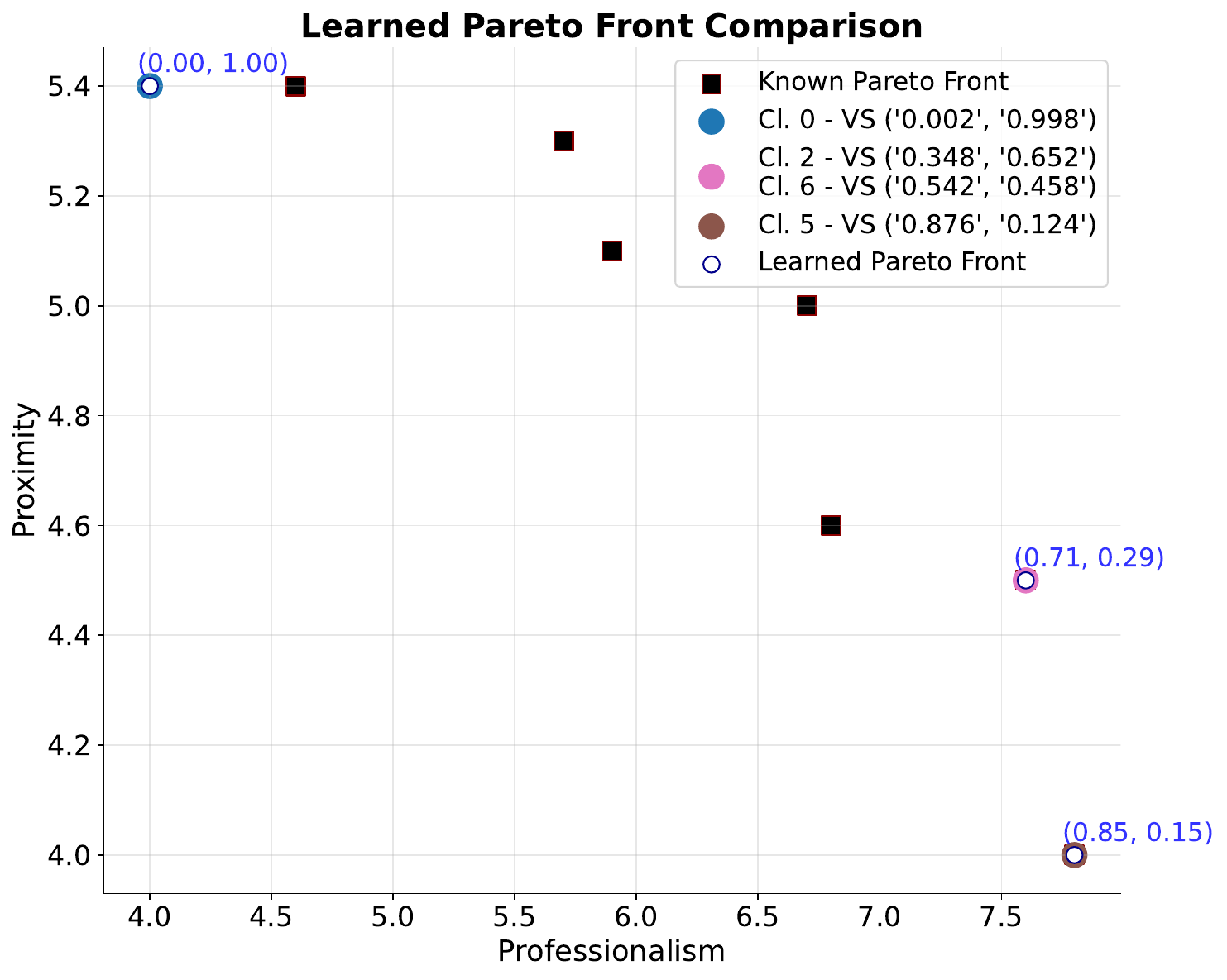}
    \includegraphics[width=0.33\linewidth,trim=0.1cm 0.2cm 0cm 1cm, clip]{media/ff/pareto/cpbmorl/seed_30.pdf}
    \includegraphics[width=0.33\linewidth,trim=0.1cm 0.2cm 0cm 1cm, clip]{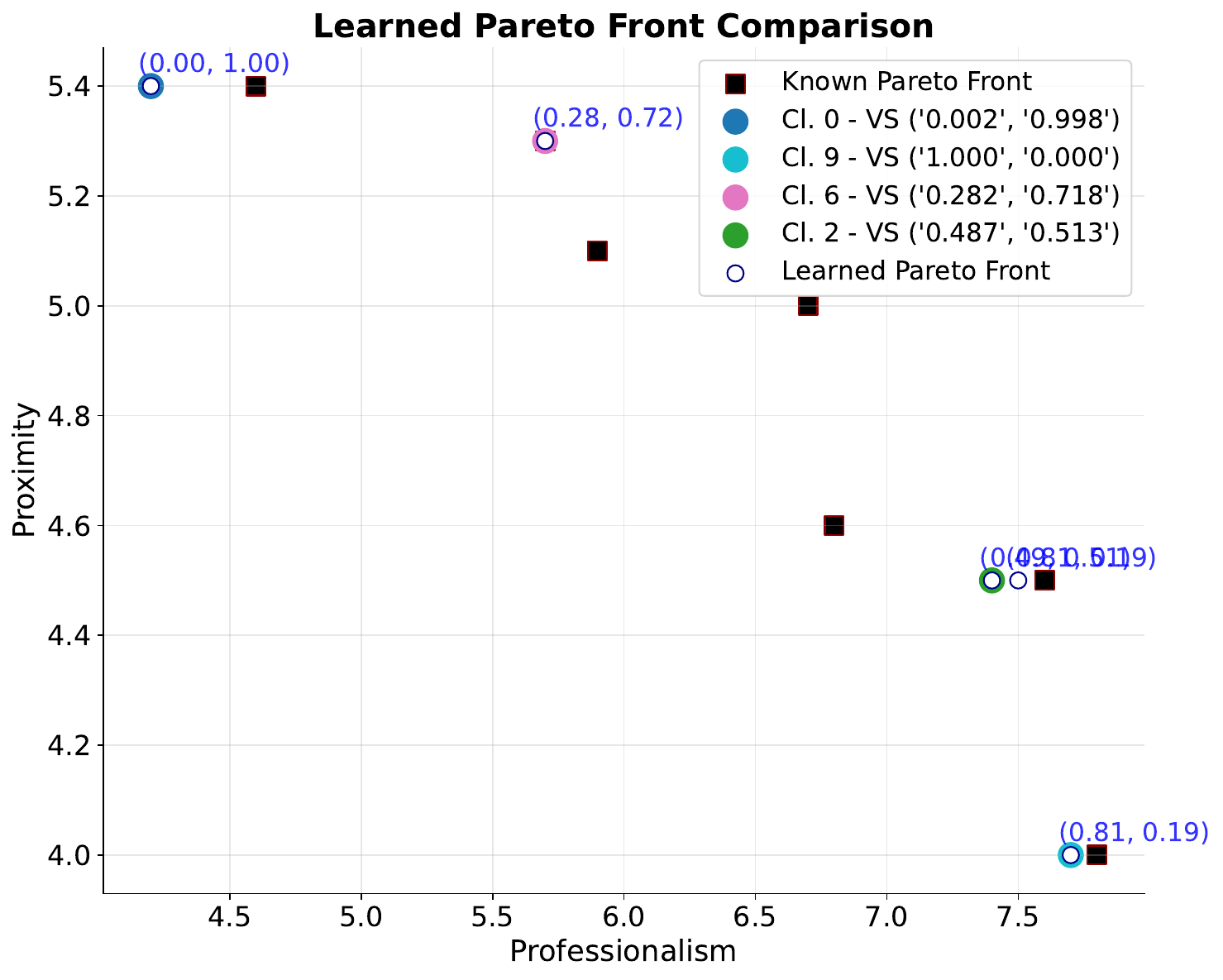}
    \includegraphics[width=0.33\linewidth,trim=0.1cm 0.2cm 0cm 1cm, clip]{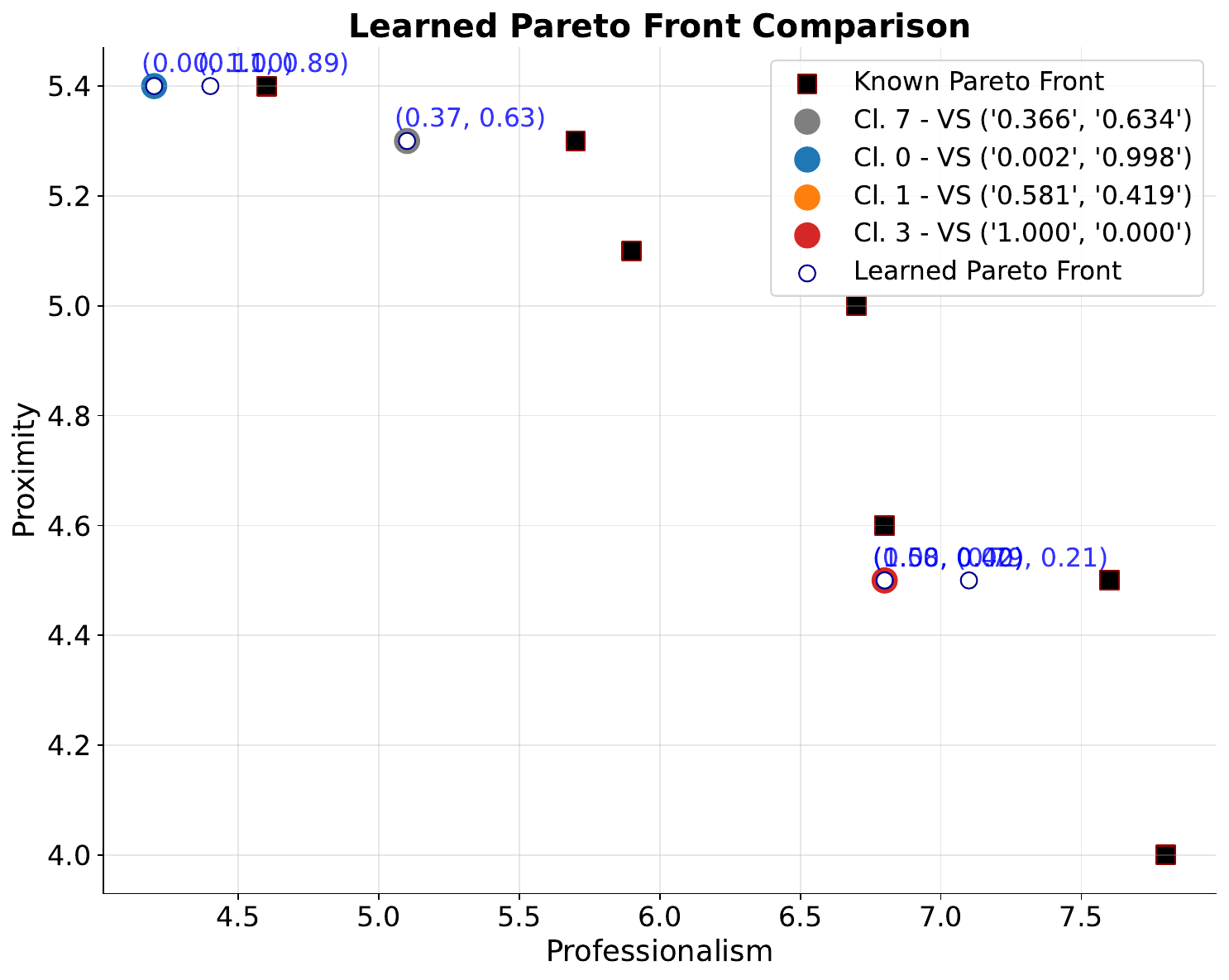}
    \includegraphics[width=0.33\linewidth,trim=0.1cm 0.2cm 0cm 1cm, clip]{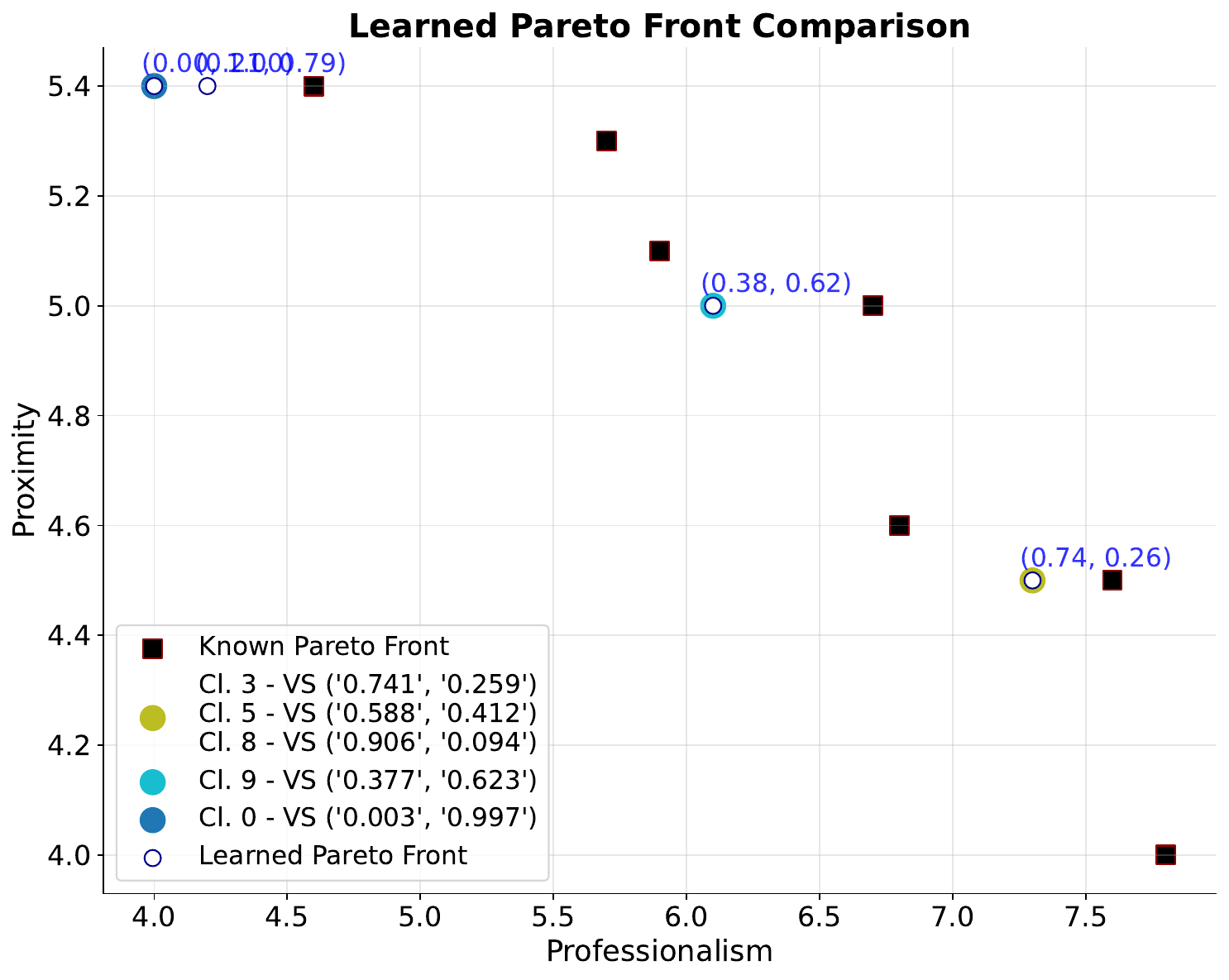}
    \includegraphics[width=0.33\linewidth,trim=0.1cm 0.2cm 0cm 1cm, clip]{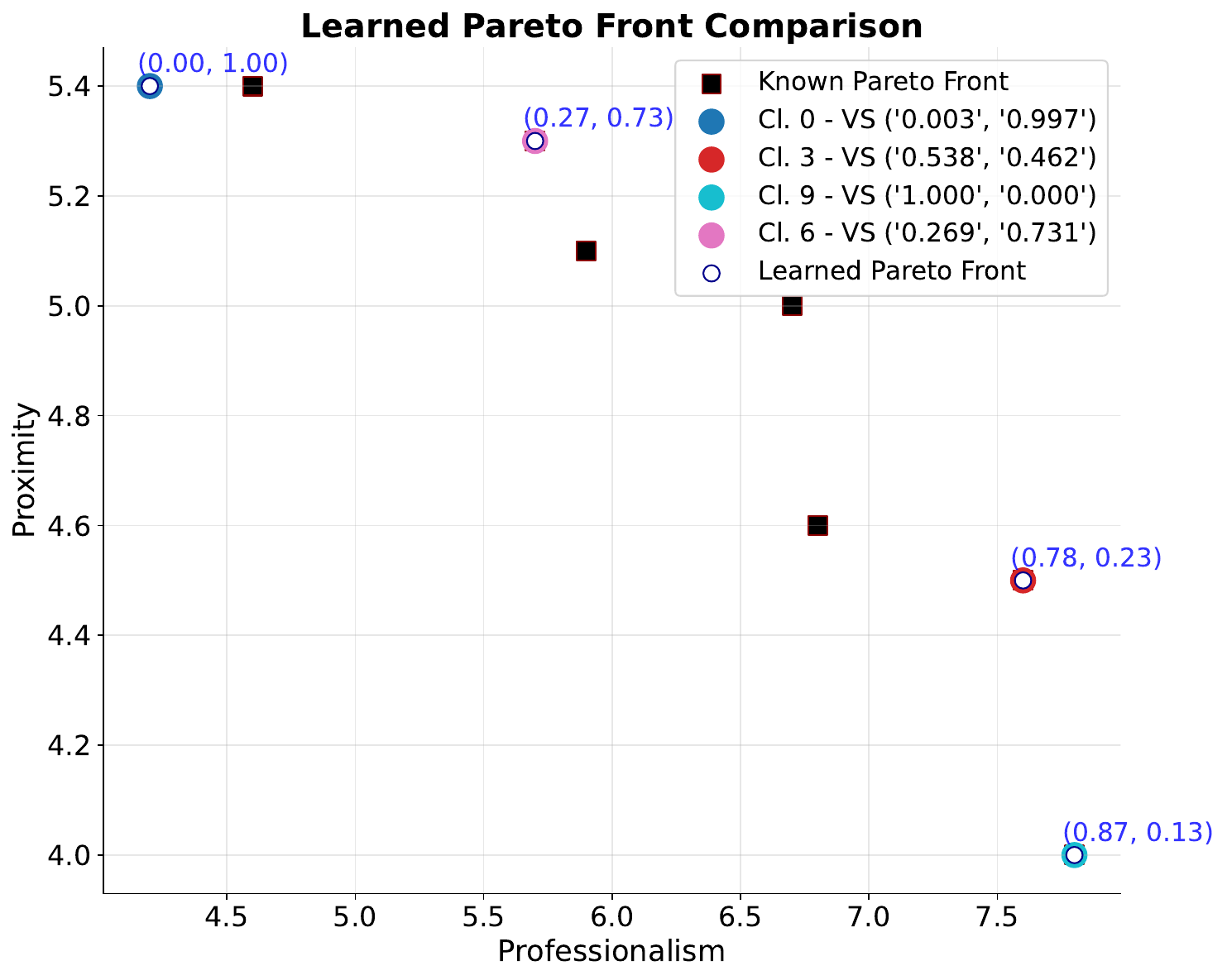}
    
    \caption{FF environment. Pareto front and clusters learned with SVSL-P with the different 10 seeds. Black squares indicate the known Pareto front of the environment in terms of the alignment with the two values. White dots depict weights which policies are in the learned front with each method. Coloured white dots indicate the value system weights identifying each learned cluster (in the legend). Note that not all of the latter are necessarily efficient. }
    \label{fig:paretoeqlffSVSLP}
\end{figure*}

\newpage
\mbox{}
\newpage
\mbox{
}
\newpage
\mbox{
}
\newpage
\mbox{
}
\newpage
\mbox{
}

\end{document}